\documentclass{article}

\usepackage{microtype}
\usepackage{graphicx}
\usepackage{subcaption}
\usepackage{booktabs} 
\usepackage[flushleft]{threeparttable}
\usepackage{multirow}

\usepackage{hyperref}

\usepackage[T1]{fontenc}
\usepackage[utf8]{inputenc}

\usepackage[accepted]{icml2025}

\usepackage{amsmath}
\usepackage{amssymb}
\usepackage{mathtools}
\usepackage{amsthm}

\usepackage[capitalize,noabbrev]{cleveref}

\DeclareMathOperator*{\argmax}{arg\,max}
\DeclareMathOperator*{\argmin}{arg\,min}

\theoremstyle{plain}

\theoremstyle{definition}

\theoremstyle{remark}

\usepackage[textsize=tiny]{todonotes}

\usepackage{xspace}
\newcommand{\hbbodkl}{\texttt{HbBoPs}\xspace}

\usepackage{xcolor}
\usepackage{xurl}

\usepackage{adjustbox}
\usepackage[nice]{nicefrac}

\icmltitlerunning{Hyperband-based Bayesian Optimization for Black-box Prompt Selection}

\begin{document}

\twocolumn[
\icmltitle{Hyperband-based Bayesian Optimization for Black-box Prompt Selection}



\icmlsetsymbol{equal}{*}

\begin{icmlauthorlist}
\icmlauthor{Lennart Schneider}{awsx}
\icmlauthor{Martin Wistuba}{aws}
\icmlauthor{Aaron Klein}{aaron}
\icmlauthor{Jacek Golebiowski}{jacek}
\icmlauthor{Giovanni Zappella}{aws}
\icmlauthor{Felice Antonio Merra}{felice}

\end{icmlauthorlist}

\icmlaffiliation{awsx}{Work done during an internship at Amazon Web Services, Berlin, Germany.}
\icmlaffiliation{aws}{Amazon Web Services, Berlin, Germany.}
\icmlaffiliation{aaron}{ScaDS.AI, University of Leipzig, Germany; work done while at Amazon.}
\icmlaffiliation{jacek}{distil labs, Berlin, Germany; work done while at Amazon.}
\icmlaffiliation{felice}{Cognism, Remote, Italy; work done while at Amazon}

\icmlcorrespondingauthor{Lennart Schneider}{lennart.sch@web.de}

\icmlkeywords{Machine Learning, ICML, black-box LLMs, prompt selection, prompt engineering, prompt optimization, Bayesian optimization, Hyperband, bandit, multi-fidelity, natural language, large language models}

\vskip 0.3in
]



\printAffiliationsAndNotice{}  

\begin{abstract}

Optimal prompt selection is crucial for maximizing large language model (LLM) performance on downstream tasks, especially in black-box settings where models are only accessible via APIs.
Black-box prompt selection is challenging due to potentially large, combinatorial search spaces, absence of gradient information, and high evaluation cost of prompts on a validation set.
We propose \hbbodkl, a novel method that combines a structural-aware deep kernel Gaussian Process with Hyperband as a multi-fidelity scheduler to efficiently select prompts.
\hbbodkl uses embeddings of instructions and few-shot exemplars, treating them as modular components within prompts.
This enhances the surrogate model's ability to predict which prompt to evaluate next in a sample-efficient manner.
Hyperband improves query-efficiency by adaptively allocating resources across different fidelity levels, reducing the number of validation instances required for evaluating prompts.
Extensive experiments across ten diverse benchmarks and three LLMs demonstrate that \hbbodkl outperforms state-of-the-art methods in both performance and efficiency.
\end{abstract}

\section{Introduction}
\label{sec:introduction}

In recent years, pre-trained auto-regressive large language models (LLMs) have demonstrated remarkable capabilities in addressing a wide range of machine learning tasks involving natural language \citep{brown2020language,liu2023pre}, such as Q\&A \citep{joshi2017triviaqa,clark2018think}, text summarization \citep{koupaee2018wikihow}, text generation~\citep{hendrycks2020measuring}, and mathematical problem-solving \citep{hendrycks2020measuring,cobbe2021training}. 
As LLMs are highly sensitive to their input \citep{zhou2022large,honovich2022instruction,lu2021fantastically,liu2023pre,ye2023compositional,wu2024prompt}, performance on these tasks relies on prompt engineering, where the input is formatted within a carefully designed prompt that may include an \emph{instruction}, a \emph{few-shot exemplar}, and additional information.

The goal of static black-box prompt optimization and selection \citep{sun2022black,chen2023instructzero,lin2023use,wu2024prompt,shi2024best} is to construct or identify a single prompt for a black-box LLM that, in expectation, performs well across all instances of a downstream task.
This process involves evaluating different prompts on a validation set and using derivative-free techniques to guide optimization or selection. 
The static black-box setting allows for offline optimization, with the resulting optimal prompt being used for the downstream task.

While much research has focused on automatically \emph{generating} new prompts \citep{sun2022black,xu2022gps,zhou2022large,chen2023instructzero,fernando2023promptbreeder,lin2023use}, there is growing interest in efficiently \emph{selecting} prompts from a predefined pool of candidates \citep{shi2024best}. 
This is because many prompt optimization techniques involve generating a large candidate pool a priori before identifying the best prompt \citep{xu2022gps,zhou2022large,fernando2023promptbreeder,prasad2022grips}.
Moreover, recent empirical findings indicate that few-shot exemplars often contribute more strongly to prompt performance than the instruction of a prompt itself, with the best results typically achieved through the optimal selection of both components \citep{wan2024joint}.

The task of black-box prompt selection is challenging.
First, the search space can be extensive, as instructions and few-shot exemplars form a combinatorial set of candidates.
Second, black-box LLMs make it impossible to directly optimize based on gradient information.
Third, evaluating a prompt is time-consuming and costly, as each evaluation involves querying the LLM on multiple validation instances of a task.
This calls for the development of selection methods that can efficiently explore the space of candidate prompts while keeping the total number of LLM calls low.


Existing methods for prompt selection in this setting include \texttt{MIPROv2} \citep{opsahl2024optimizing}, \texttt{EASE} \citep{wu2024prompt}, \texttt{TRIPLE-SH}, and \texttt{TRIPLE-GSE} \citep{shi2024best}.
We identify the following limitations among these approaches:
(1) Except for \texttt{MIPROv2}, these methods are not explicitly designed to address the problem of \emph{jointly} selecting instructions and few-shot exemplars.
Specifically, \texttt{EASE} primarily focuses on exemplar selection, while \texttt{TRIPLE-SH} and \texttt{TRIPLE-GSE} focus on instruction selection.
Although \texttt{EASE} can be applied to joint selection by treating the entire prompt as an unstructured block of text, it does not exploit the compositional structure of prompts.
(2) No method is both \emph{sample-efficient} (allowing for evaluating fewer prompts by relying on a surrogate model) and \emph{query-efficient} (reducing the total number of LLM calls by not evaluating prompts on all available validation instances).

In this work, we propose \hbbodkl (\textbf{H}yper\textbf{b}and-based \textbf{B}ayesian \textbf{o}ptimization for black-box \textbf{P}rompt \textbf{s}election) addressing these limitations.
Our main contributions are the following:
(1) We introduce a structural-aware deep kernel Gaussian Process (GP) that learns a low-dimensional prompt representation from separate embeddings of instructions and few-shot exemplars in an end-to-end fashion to predict prompt performance.
(2) We adopt Hyperband \citep{li2018hyperband} as a \emph{multi-fidelity} scheduler for prompt selection that governs the number of validation instances prompts are evaluated on.
(3) We introduce a novel method, \hbbodkl, that relies on our structural-aware deep kernel GP to make a Bayesian Optimization (BO) proposal within Hyperband and as a result is both sample- and query-efficient.
(4) We compare \hbbodkl against four baselines and four state-of-the-art methods across ten benchmarks and three LLMs, demonstrating that \hbbodkl performs best in identifying a well-performing prompt after a given budget of total LLM calls but also exhibits strongest anytime performance during the selection process.
(5) We perform an ablation study of the components of \hbbodkl gaining insight into their inner workings and further demonstrate its robustness to the choice of the encoder model used to obtain embeddings.

\section{Problem Statement}
\label{sec:problem_statement}
Let $\mathcal{I} = \{i_{1}, \ldots, i_{l}\}$ denote a finite set of instructions (task descriptions) and $\mathcal{E} = \{e_{1}, \ldots, e_{m}\}$ a finite set of few-shot exemplars.
Note that by \emph{exemplar} we refer to an ordered tuple of a given number of input-output examples of a task.
Let $\mathcal{P} = \mathcal{I} \times \mathcal{E}$ be the set of prompts that are generated by combining each $i \in \mathcal{I}$ with each $e \in \mathcal{E}$.

Instructions can be generated either manually by experts or automatically by LLM-based methods, e.g., Automatic Prompt Engineering (APE; \citealt{zhou2022large}).
Few-shot exemplars can be generated by selecting different input-output instances from a training set of the task.

A prompt $p \in \mathcal{P}$ is instantiated by combining it with a given task input or query $x \in \mathcal{X}$ for which the LLM, $h: (\mathcal{P} \times \mathcal{X}) \rightarrow \mathcal{Y}, h([p,x]) \mapsto \hat{y}$, produces an output $\hat{y} \in \mathcal{Y}$.
We will use $h_{p}(x)$ to denote $h([p,x])$ as a shorthand.

We make no assumptions regarding the nature of the LLM and treat it as a black-box.
The LLM returns output given input without any additional information, i.e., no access to model parameters, gradients, or token probabilities.

Having access to a validation set $\{(x_i, y_i)\}_{i=1}^{n_{\mathrm{valid}}}$, evaluating a prompt is performed by comparing the ground truth output $y_i$ to the output $\hat{y}_i = h_p(x_i)$ generated by the LLM based on a pointwise loss function $l: \mathcal{Y} \times \mathcal{Y} \rightarrow \mathbb{R}, (y, \hat{y}) \mapsto l(y, \hat{y})$.
This loss function quantifies how close the output generated by the LLM is to the ground truth.
For example, a loss function based on the widely used exact match \citep{chang2024survey} scoring function is given by:

\begin{equation}
\label{eq:em_loss}
l(y, \hat{y}) =
\begin{cases} 0 & \text{if } y = \hat{y} \\ 1 & \text{otherwise}. \end{cases}
\end{equation}

Our task is to identify a single prompt $p \in \mathcal{P}$ that is optimal with respect to the loss in expectation:
\begin{equation}
\label{eq:problem_theory}
\argmin_{p \in \mathcal{P}} \mathbb{E}_{(x, y) \sim \mathbb{P}_{xy}} \left[ l(y, h_{p}(x)) \right].
\end{equation}
Here, the expectation is taken over all input-output instances $(x, y)$.
In practice, however, Equation~\eqref{eq:problem_theory} can only be approximated based on the validation instances available:
\begin{equation}
\label{eq:black_box_practice}
f(p) \coloneqq \frac{1}{n_{\mathrm{valid}}} \sum_{i=1}^{n_{\mathrm{valid}}} l(y_i, h_{p}(x_i)).
\end{equation}

We refer to this setting as the \emph{static} setting, as we are searching offline for a single optimal prompt on the target downstream task.
Note that due to the non-deterministic nature of LLMs (depending on temperature), $f$ itself can in general only be observed with noise.

We will denote by 
\begin{equation}
\label{eq:observed_error}
v = f(p) + \epsilon,~\epsilon \sim \mathcal{N}(0, \sigma^2)
\end{equation}
the observed validation error of the LLM configured to use prompt $p$ \footnote{We model the observation noise as homoscedastic Gaussian for analytical convenience.}.
$f$ is a black-box as the LLM is a black-box and no analytic description or gradient information is available.

The core challenge of static black-box prompt selection is balancing exploration (searching the prompt space $\mathcal{P}$) and efficiency (minimizing costly LLM evaluations).
Our goal is to identify the best prompt using as few LLM calls for evaluation as possible.

\section{Method}
\label{sec:method}


We want to learn a surrogate model of the black-box function $f$ in Equation~\eqref{eq:black_box_practice} that predicts the validation error of prompts on a downstream task based on observed data collected during optimization.
This surrogate is used to predict the validation error of the unevaluated prompts during the optimization process.
We first describe how GPs can serve as surrogate models in black-box prompt selection (Section~\ref{sec:gp}), but highlight the limitations of vanilla GPs on raw, high-dimensional prompt embeddings.
To address this, we introduce a structural-aware deep kernel GP (Section~\ref{sec:structural_aware_dkl}) that relies on structural information of prompts via separate embeddings of their building blocks.
We then adopt Hyperband \citep{li2018hyperband} for query-efficient multi-fidelity prompt selection (Section~\ref{sec:hb}) and integrate our structural-aware deep kernel GP via a BO proposal, which results in our final \hbbodkl algorithm (Section~\ref{sec:hbbodkl}).

\subsection{Gaussian Process as Surrogate Model}\label{sec:gp}

To learn a surrogate model, we collect, at each optimization step $t$, a set of design data $\mathcal{D}_t := \{(p_j, v_j)\}_{j = 1}^{t}$, where each tuple consists of a prompt $p_j \in \mathcal{P}$ and its corresponding validation error $v_j$, as defined in Equation~\eqref{eq:observed_error}, recorded at the $j$-th previous step.
This design data captures the history of prompts evaluated and their observed performance throughout the sequential optimization process.
To learn a model that maps prompts to their validation errors, we embed each prompt into a $d$-dimensional numeric space, making use of pre-trained language encoders.
Let $enc: \mathcal{P} \rightarrow \mathbb{R}^d, p \mapsto \mathbf{z}$ be the encoding function.
We then augment the design data, $\mathcal{D}_t = \{(p_j, \mathbf{z}_j, v_j)\}_{j=1}^{t}$, where $\mathbf{z}_j$ is the embedding of prompt $p_j$.
Since we are concerned with black-box prompt selection, it is natural to use embeddings as feature representations of prompts.
Notably, recent work \citep{tang2024llm} has shown that (LLM) embeddings can in general serve as effective features for high-dimensional regression tasks, even in domains where the input data is not textual.

We want to use a GP as a surrogate model since it allows for flexible probabilistic modeling of black-box functions by returning a point estimate and well-calibrated uncertainty estimates in the form of a Gaussian posterior predictive distribution \citep{williams2006gaussian}. 
In the following, we assume a GP prior over $f$ in the $d$-dimensional space of embedded prompts, $f(\mathbf{z}) \sim \mathcal{GP}(m, k)$; $\mathbf{f} \sim \mathcal{N}(m(\mathbf{Z}), k(\mathbf{Z}, \mathbf{Z}|\mathbf{\theta}))$, where 
$m$ is the prior mean function, usually set to zero,
$k$ is the covariance function depending on kernel parameters $\mathbf{\theta}$, and
$\mathbf{Z}$ is a matrix of prompt embeddings. 

Given the design data $\mathcal{D}_t$ and new prompts $\mathbf{p}_\star$ with their embeddings $\mathbf{Z}_\star$, the function $\mathbf{f}_\star$ is modeled as a random variable that is jointly Gaussian distributed with all previously observed validation errors $\mathbf{v} = (v_1, \ldots, v_t)$.

In short, this can be written as
\begin{equation*}
\begin{bmatrix}
\mathbf{v} \\
\mathbf{f}_\star
\end{bmatrix} 
\sim 
\mathcal{N}\left(
\begin{bmatrix} m(\mathbf{Z})\ m(\mathbf{Z}_\star)\end{bmatrix}
, \begin{bmatrix}
\mathbf{K}_{t} & \mathbf{K}_\star\\
\mathbf{K}^\intercal_\star & \mathbf{K}_{\star\star},
\end{bmatrix}\right),
\end{equation*}
where $\mathbf{K}_t = k(\mathbf{Z},\mathbf{Z}|\mathbf{\theta}) + \sigma^2\mathbf{I}_t$, $\mathbf{K}_\star = k(\mathbf{Z},\mathbf{Z}_\star|\mathbf{\theta})$, and  $\mathbf{K}_{\star\star} = k(\mathbf{Z}_\star,\mathbf{Z}_\star|\mathbf{\theta})$ are the kernel matrices.

The posterior predictive distribution under the (zero mean) GP is obtained as
\begin{equation}
\label{eq:posterior_predictive}
    \begin{split}
        \mathbb{E}[\mathbf{f}_\star|\mathbf{Z},\mathbf{v},\mathbf{Z}_\star] &= \mathbf{K}^\intercal_\star (\mathbf{K}_t)^{-1} \mathbf{v}, \\
        \mathrm{cov}[\mathbf{f}_\star|\mathbf{Z},\mathbf{Z}_\star] &= \mathbf{K}_{\star\star} - \mathbf{K}^\intercal_\star (\mathbf{K}_t)^{-1} \mathbf{K}_\star,
    \end{split}
\end{equation}
where common choices for the kernel function $k$ are given by the squared exponential kernel or variations of the Matérn kernel \citep[see, e.g.,][Chapter~4]{williams2006gaussian}.

At this point, we could proceed and train a vanilla GP as outlined above on the $d$-dimensional space of embedded prompts.
However, as stated in many previous works on BO~\citep{kandasamy2015high,wang2016bayesian,gardner2017discovering,eriksson2019scalable,eriksson2021high}, GPs struggle with high-dimensional input such as that found in our design data $\mathcal{D}_t$, e.g., the dimensionality of BERT's \citep{devlin-etal-2019-bert} [CLS] token embedding is already 768.
In general, dimensionality reduction techniques such as principal component analysis (PCA) or random projections could be used.
However, such techniques are unsupervised and will not yield a lower-dimensional representation that is aligned with the downstream performance of prompts (see also Figure~\ref{fig:feature_comparison} in Appendix~\ref{app:dkl} for an illustration). 
Additionally, using a single embedding of the whole prompt treated as a block of text ignores that the prompt is composed of different building blocks with distinct structural information.
Below, we present our solution.

\subsection{Structural-aware Deep Kernel}\label{sec:structural_aware_dkl}
To learn a lower-dimensional representation of the embedded prompts aligned with the downstream task, we propose to use a deep kernel \citep{pmlr-v51-wilson16} within the GP.
We design a feature extractor, $\phi: \mathbb{R}^d \rightarrow \mathbb{R}^p$, $p \ll d$ to learn a flexible kernel transformation function $k(\phi(\mathbf{z}, \mathbf{w}),\phi(\mathbf{z}', \mathbf{w})|\mathbf{\theta})$, where $\mathbf{\theta}$ and $\mathbf{w}$ are the parameters of the kernel and respectively the extractor. 

Given that prompts consist of two distinct components (instructions and few-shot exemplars), we hypothesize that embedding these components separately can improve the deep kernel GP's (DK-GP) ability to use both structural and semantic differences effectively.
Instructions typically exhibit consistent patterns across prompts, whereas few-shot exemplars are more variable due to diverse input-output pairs and flexible ordering.
To address this, we propose learning a structural-aware latent representation of prompts.
Our approach involves embedding the instructions $i \in \mathcal{I}$ and few-shot exemplars $e \in \mathcal{E}$ separately.
Each component embedding is processed through distinct feed-forward neural networks, each consisting of two fully connected layers with ReLU activations, as defined below:

\begin{footnotesize}
\begin{flushleft}
$\phi_{\text{enc}(\cdot)}:$ \\
\vspace{0.5em}
\texttt{Lin(d, 64)} $\rightarrow$ \texttt{ReLU()} $\rightarrow$ \texttt{Lin(64, 32)} $\rightarrow$ \texttt{ReLU()}
\end{flushleft}
\end{footnotesize}

After processing the instructions and exemplars independently, we concatenate their outputs and input the combined representation into another feed-forward neural network to learn a joint latent representation:

\begin{footnotesize}
\begin{flushleft}
$\phi_{\left( \phi_{\text{enc}(i)}, \phi_{\text{enc}(e)} \right)}:$ \\
\vspace{0.5em}
\texttt{Lin(32 }$\!\cdot\!$\texttt{ 2, 32)} $\rightarrow$ \texttt{ReLU()} $\rightarrow$ \texttt{Lin(32, 10)}
\end{flushleft}
\end{footnotesize}

During training the GP, we obtain both the optimal kernel parameters and the parameters of the neural network feature extractor $\phi$ by optimizing the log marginal likelihood criterion (up to constants):
\begin{equation}
\label{eq:mll}
    \hat{\mathbf{\theta}}, \hat{\mathbf{w}} = \argmax_{\mathbf{\theta},\mathbf{w}} - \mathbf{v}^\intercal \mathbf{K}_{t}(\mathbf{\theta},\mathbf{w})^{-1} \mathbf{v} - \log |\mathbf{K}_{t}(\mathbf{\theta},\mathbf{w})|
\end{equation}
An additional advantage of using a deep kernel GP is its ability to model non-stationary functions, as the feature extractor learns input-dependent transformations that allow the surrogate to capture varying levels of smoothness across the input space, which standard stationary kernels in vanilla GPs cannot accomplish \citep{li2024bnn}.
Before illustrating how we use our structural-aware DK-GP during optimization to achieve sample-efficiency via a BO proposal, we explain how we ensure query-efficiency via Hyperband.

\subsection{Hyperband for Multi-Fidelity Scheduling}\label{sec:hb}
To improve query-efficiency identified as one of the key limitations of existing works, we want to terminate the evaluation of poor-performing prompts early, saving cost during the evaluation process.
Similarly to \texttt{TRIPLE} \citep{shi2024best}, we model the number of validation instances prompts are evaluated on as a fidelity parameter.
Full-fidelity methods evaluate prompts on all validation instances, while multi-fidelity methods adaptively schedule evaluations on varying numbers of instances.
\texttt{TRIPLE} implements Successive Halving (SH; \citealt{karnin2013almost}) as a multi-fidelity scheduler.
In contrast, we use Hyperband (HB; \citealt{li2018hyperband}) as it will generally evaluate fewer prompts and hedges against a poorly configured SH as explained below.

Given a total budget of $B$ LLM calls to evaluate prompts on validation instances, SH allocates a budget of $b = B / (|\mathcal{P}| \log_2{(|\mathcal{P}|)})$ to each prompt (see details in Appendix~\ref{app:mf}).
After having evaluated the prompts on $b$ validation instances, the lower half of bad performing prompts is discarded, and the process repeats, doubling the number of calls for the remaining prompts in the next stage, until a single prompt remains. 

This strategy is affected by the ``budget vs. number of configurations'' dilemma \citep{li2018hyperband}, since, at the beginning of the algorithm, it is not clear if we should evaluate many (by default all) prompts on few instances (good exploration but noisy performance estimates) or few prompts on many instances (less exploration but accurate performance estimates).
When many prompts need to be evaluated with a limited total budget, SH's initial budget is low, which risks discarding a prompt based on noisy performance estimation.
HB in contrast hedges against a poor choice of the number of starting prompts and their budget by repeatedly running SH in different brackets with different numbers of starting prompts and starting budgets.
This results in HB being robust under various scenarios without knowing the optimal resource allocation, making it ideal for prompt selection.

To tailor HB to prompt selection, we make the following design decisions (for an ablation study, see Appendix~\ref{app:additional_results}):
(1) We extend previous evaluations when advancing stages within a bracket, ensuring validation instances of higher stages subsume those of lower stages, and (2) return the prompt with the lowest validation error among those evaluated on the entire validation set.

Finally, we combine HB with our structural-aware DK-GP by employing a sequential BO proposal mechanism for candidate prompts in each bracket, which we outline next.

\begin{algorithm}[!t]
    \caption{\hbbodkl}
    \label{alg:hyperband_modified_sequential}
\begin{algorithmic}
\INPUT $n_{\mathrm{valid}}$, $b_{\min}$ ({\tiny{lower limit to  \#validation instances}}), $\eta$ ({\tiny{halving parameter}})
    \STATE $r = n_{\mathrm{valid}} / b_{\min}$
    \STATE $s_{\max} = \lfloor\log_{\eta}(r)\rfloor$
    \STATE $B = (s_{\max} + 1)n_{\mathrm{valid}}$
    \FOR{$s \in \{s_{\max}, s_{\max} - 1, \ldots, 0\}$}
        \STATE $n = \left\lceil\frac{B}{n_{\mathrm{valid}}}\frac{\eta^s}{(s+1)}\right\rceil$
        \STATE $b = n_{\mathrm{valid}}\eta^{-s}$
        \STATE $P = \{\}$, $V = \{\}$
        \FOR{$j \in \{0, \ldots, n-1\}$}
            \STATE \colorbox{gray!30}{$p = \text{get\_prompt}()$}
            \STATE $v = \text{get\_validation\_error}(p, b)$
            \STATE $P \leftarrow P \cup \{p\}$, $V \leftarrow V \cup \{v\}$
        \ENDFOR
        \STATE $P = \text{top\_k}(P, V, \lfloor n/\eta \rfloor)$
        \FOR{$i \in \{1, \ldots, s\}$}
            \STATE $n_i = \lfloor n\eta^{-i} \rfloor$
            \STATE $b_i = b\eta^i$
            \STATE $V = \{\text{get\_validation\_error}(p, b_i): p \in P\}$
            \STATE $P = \text{top\_k}(P, V, \lfloor n_i/\eta \rfloor)$
        \ENDFOR
    \ENDFOR
    \OUTPUT Prompt with the lowest validation error evaluated on the whole validation set
\end{algorithmic}
\end{algorithm}

\subsection{\hbbodkl}\label{sec:hbbodkl}

\hbbodkl combines HB with our structural-aware DK-GP.
While the vanilla HB algorithm for prompt selection samples prompts uniformly at random, \hbbodkl replaces the random proposal mechanism of HB with a sequential BO proposal (highlighted in gray in Algorithm~\ref{alg:hyperband_modified_sequential}).
This is similar to the approach proposed by \citet{falkner2018bohb} for hyperparameter optimization.
During the execution of HB, \hbbodkl trains the GP on a subset of the design data ${\mathcal{D}_{t}}_{|b}$ for a given fidelity-level $b$.
We use the highest fidelity $b$ for which ``enough'' observations are available.
This design decision stems from the observation that validation errors are estimated more accurately with more instances (see Appendices~\ref{app:generalization} and \ref{app:mf}).
Importantly, we train the GP entirely online during the selection process and are not relying on a pre-trained surrogate model.

After training the GP on ${\mathcal{D}_{t}}_{|b}$, we select the next candidate prompt $p_{t+1}$ by maximizing the Expected Improvement (EI; \citealt{kushner1964bo,mockus1978,jones1998efficient}) acquisition function:
\begin{equation}
    \begin{gathered}
    \alpha_{\mathrm{EI}}(p|{\mathcal{D}_{t}}_{|b}) \coloneqq \mathbb{E}[\max\{v_{\min,b} - f(\mathbf{z}_p), 0\}] \\
    p_{t+1} = \argmax_{p \in \mathcal{P}} \alpha_{\mathrm{EI}}(p|{\mathcal{D}_{t}}_{|b}),
    \end{gathered}
\end{equation}
In words, given the incumbent (the best-performing prompt evaluated at the highest fidelity level so far) and its validation error $v_{\min,b}$ at fidelity level $b$, the EI quantifies the expected improvement of a candidate prompt over the incumbent (considering only actual improvement due to the $\max$ operator), based on the GP's posterior predictive distribution (Equation~\eqref{eq:posterior_predictive}).
Since our search space is given by a finite set of candidate prompts, we can evaluate the EI exhaustively for all candidate prompts.

\begin{table*}[!t]
\centering
\caption{Overview of baselines, competitors and our \hbbodkl in the static black-box prompt selection setting.}
\label{tab:method_comparison}
\begin{adjustbox}{width=\textwidth,center}
\begin{tabular}{lcccccc}
\toprule
\textbf{Method} & 
\multirow{2}{*}{\begin{tabular}[c]{@{}c@{}}\textbf{Fidelity} \\ \textbf{Level}\end{tabular}} & 
\multicolumn{2}{c}{\textbf{Efficiency}} & 
\multirow{2}{*}{\begin{tabular}[c]{@{}c@{}}\textbf{Surrogate} \\ \textbf{Model}\end{tabular}} &
\multirow{2}{*}{\begin{tabular}[c]{@{}c@{}}\textbf{Bandit} \\ \textbf{Algorithm}\end{tabular}} &
\multirow{2}{*}{\begin{tabular}[c]{@{}c@{}}\textbf{Prompt} \\ \textbf{Representation}\end{tabular}}\\
\cmidrule(lr){3-4}
 & & sample & query \\
\midrule
\texttt{RS} & Full & - &  - & - & - & $p$ \\
Vanilla \texttt{BO} & Full & \checkmark & - & vanilla GP &- & $enc(p)$ \\
\texttt{HDBO} & Full & \checkmark &  - & GP \citep{hvarfner2024vanilla} &- & $enc(p)$ \\
\texttt{BOPCA} & Full & \checkmark &  - & vanilla GP & - & $\text{PCA}(enc(p))$ \citep{zhang2024language} \\ \midrule
 \texttt{EASE} \cite{wu2024prompt} & Full & \checkmark &  - & NN & NUCB & $enc(p)$ \\
\texttt{MIPROv2} \cite{opsahl2024optimizing} & Full & \checkmark &  - & TPE & - & $\text{ID}_i\,\text{ID}_e$ \\
\texttt{TRIPLE-SH} \cite{shi2024best}& Multi & - & \checkmark &  - & SH & $p$ \\
\texttt{TRIPLE-GSE} \cite{shi2024best} & Multi & - & \checkmark  & LM/GLM & GSE & $enc(p)$ \\ \midrule
\hbbodkl (ours) & Multi &  \checkmark & \checkmark & structural-aware DK-GP & HB & $enc(i), enc(e)$ \\
\bottomrule
\end{tabular}
\end{adjustbox}
\vspace{-1em}
\end{table*}

\section{Experimental Setup}
\label{sec:experimental_setup}

\subsection{Benchmark Tasks}
We benchmark \hbbodkl on ten tasks commonly used for LLM evaluation \citep{zhou2022large,lin2023use,chen2023instructzero,wu2024prompt,shi2024best}.
\emph{AI2’s Reasoning Challenge (ARC)}~\citep{clark2018think}: multiple-choice Q\&A problems;
\emph{Grade School Math 8K}~\citep{cobbe2021training}: math problems taking between two and eight steps to solve;
\emph{Eight Tasks from the BBII subset of the BIG-bench and instruction induction benchmarks} \citep{srivastava2022beyond,honovich2022instruction} used in \citet{zhou2022large,wu2024prompt,shi2024best}: antonyms, larger animal, negation, second word letter, sentiment, object counting, orthography starts with, and word unscrambling.
Task statistics are reported in Table~\ref{tab:tasks} in Appendix~\ref{app:experiments}.

\subsection{Methods}
We compare \hbbodkl against full-fidelity and multi-fidelity methods described in Table~\ref{tab:method_comparison}.
Additional details on the methods are reported in Section~\ref{subsec:static-black-box-methods} and Appendix~\ref{app:experiments}.
\texttt{RS} is a simple random search.
All methods that rely on embeddings of prompts use BERT's [CLS] token embedding to ensure fair comparison.
All full-fidelity BO methods (vanilla BO, \texttt{HDBO}, \texttt{BOPCA}) use an ARD Matérn $\nicefrac{5}{2}$ kernel and Expected Improvement as acquisition function and normalize inputs to the unit cube and standardize outputs to have zero mean and unit variance.
\texttt{HDBO} is a simple but well-performing high-dimensional BO algorithm using adjusted priors on GP kernel and likelihood parameters, as described in \citet{hvarfner2024vanilla}.
\texttt{BOPCA} uses a ten component PCA inspired by \citet{zhang2024language}.
We run \texttt{MIPROv2}, NUCB \citep{pmlr-v119-zhou20a} as used by \texttt{EASE} ($\nu = 0.1$), \texttt{TRIPLE-SH} and \texttt{TRIPLE-GSE} as implemented in their publicly available code bases.
All full-fidelity methods use the same initial design of ten prompts sampled uniformly at random.
\hbbodkl uses an ARD Matérn $\nicefrac{5}{2}$ kernel, normalizes inputs to the unit cube and standardizes outputs.
We always train the DK-GP on the highest fidelity level for which at least four observations are available.
To optimize the log marginal likelihood in Equation~\eqref{eq:mll}, we use AdamW \citep{loshchilov2017decoupled} with learning rate $=0.01$, maximum number of epochs $=3000$, and early termination with a patience $=10$.
Within the HB schedule, we use a lower limit on the number of validation instances $b_{\min} = 10$ and a halving parameter $\eta = 2.0$.

\subsection{Experimental Protocol}
For each task, we generate a search space $\mathcal{P}$ of candidate prompts by combining five task-specific instructions with 50 few-shot exemplars.
Instructions are generated using APE's forward mode \citep{zhou2022large}, where Claude 3 Sonnet \citep{TheC3} generates instructions based on ten input-output samples from each task's training set.
For exemplars, we sample 25 sets of \emph{five} input-output instances from the training set of each task, then permute each set twice to create 50 exemplar tuples, allowing assessment of example ordering effects.
We fix the five-shot setting throughout our experiments to ensure consistency with prior work \citep{wu2024prompt}.
The final $|\mathcal{P}| = 250$ prompts are formed by the Cartesian product of instructions and exemplars.

For LLMs, we use Claude 3 Haiku \citep{TheC3}, LLAMA3 8B Instruct \citep{dubey2024llama}, and Mistral 7B Instruct \citep{jiang2023mistral}.

For each benchmark scenario (a given task and LLM), we run each method for a total budget of 25 full-fidelity evaluations (i.e., being allowed as many LLM calls as 25 prompts evaluated on all validation instances require for a given task) to mimic a budget-constrained scenario.
We use the number of LLM calls as our cost metric, rather than actual monetary cost, since LLM calls are a directly interpretable and model-agnostic measure of cost that allows for aggregating over different benchmark scenarios.
We repeat each method run 30 times on each benchmark scenario. 
We evaluate prompts using the loss function described in Equation~\eqref{eq:em_loss} which is based on the exact match scoring function.

\begin{figure*}[!t]
    \centering
    \begin{subfigure}{.5\textwidth}
        \includegraphics[width=\linewidth]{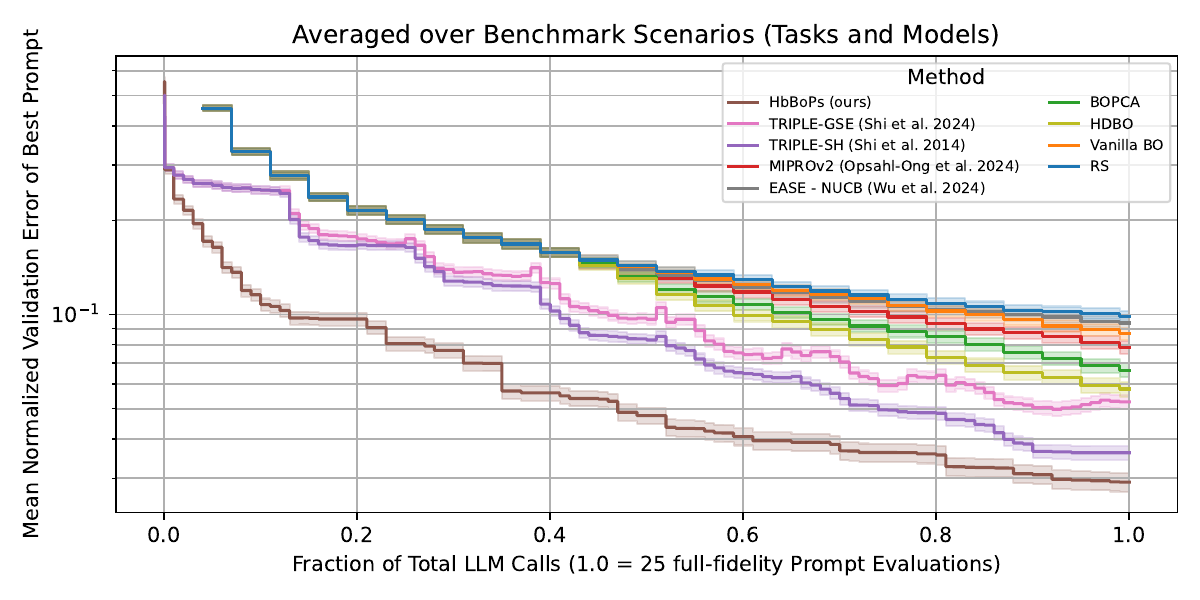}
        \caption{Validation}
        \label{fig:main_overall_valid}
        \vspace{-1em}
    \end{subfigure}%
    \begin{subfigure}{.5\textwidth}
        \centering
        \includegraphics[width=\linewidth]{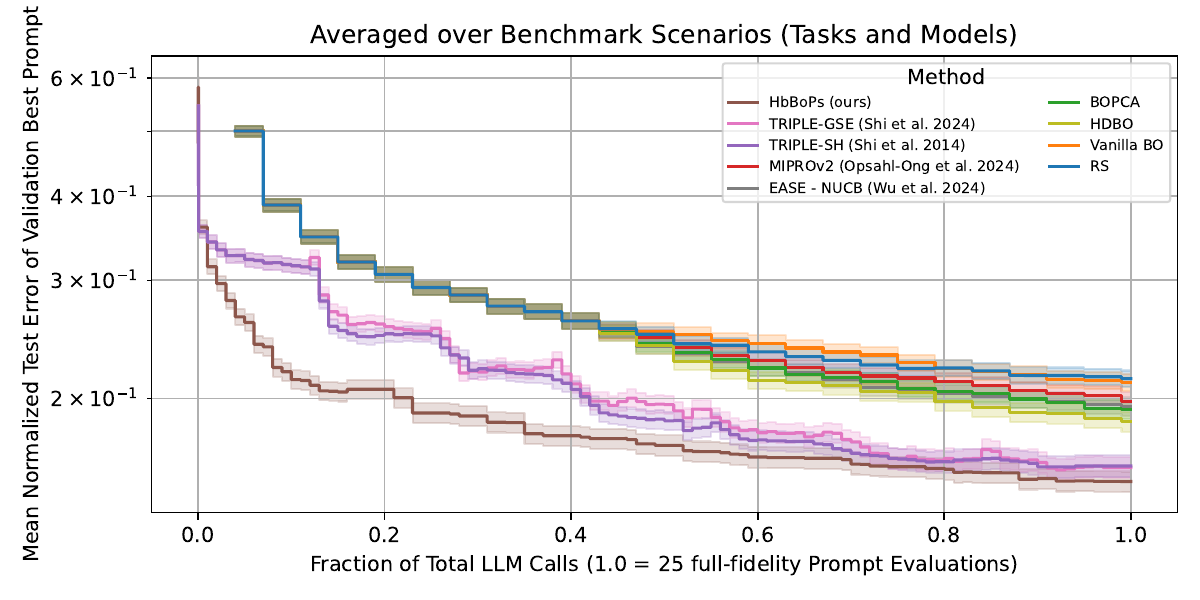}
        \caption{Test}
        \label{fig:main_overall_test}
        \vspace{-1em}
    \end{subfigure}%
    \caption{Normalized error (log scale) of the best prompt per method, averaged over benchmarks. Lower is better. Ribbons represent SE.} 
    \label{fig:main_overall}
    \vspace{-1em}
\end{figure*}

\section{Results}\label{sec:results}
We report the validation and test errors computed for the best prompt identified by each method given a specific budget.
For instance, given \emph{GSM8k} with a validation set of 1319 instances, a budget of $0.25$ means that we report the results of the methods after performing $\lceil0.25\cdot25\cdot1319\rceil=8244$ LLM calls.
Therefore, full-fidelity methods always start after having executed a fraction of $\nicefrac{1}{25}$ total LLM calls.

\subsection{Analysis of Overall Performance}\label{sec:main_results}
We start by analyzing the overall performance of the methods averaged across all benchmark tasks and LLMs.
To allow for averaging results, we normalize validation and test errors for each benchmark scenario by the performance of the worst and best prompt.
Figure~\ref{fig:main_overall} visualizes the results.
We observe that \hbbodkl outperforms all full-fidelity and multi-fidelity methods, particularly in terms of anytime performance on both the validation and test set. 

Beginning with an analysis of test error at full budget (i.e., a fraction of LLM calls equal to 1.0), we can see that our \hbbodkl on average outperforms all full-fidelity and multi-fidelity approaches with an average normalized test error of 0.150.
In detail, we observe that all full-fidelity methods surpass the \texttt{RS} baseline (0.214) with the following errors: Vanilla \texttt{BO} (0.211), \texttt{MIPROv2} (0.198), \texttt{EASE} (0.195), \texttt{BOPCA} (0.192), and \texttt{HDBO} (0.185).
However, they all have higher error values than \hbbodkl (0.150).
Additionally, \hbbodkl also outperforms all multi-fidelity methods.
Although both \texttt{TRIPLE-GSE} (0.158) and \texttt{TRIPLE-SH} (0.159) exhibit superior performance compared to their best-in-class full-fidelity counterpart, i.e., \texttt{HDBO} (0.185), they on average have identified prompts that yield error values higher than the ones obtained for \hbbodkl's prompts.

Looking at the anytime performance with a more limited budget, e.g., a fraction of 0.25 LLM calls, we can confirm \hbbodkl's improvements over the baselines.
Indeed, \hbbodkl on average outperforms \texttt{HDBO}, the best full-fidelity method, by approximately 35\%  and \texttt{TRIPLE-SH}, the best multi-fidelity method, by 24\%.
For additional statistical analyses, we refer to Appendix~\ref{app:additional_results}.

\subsection{Analysis of the Performance for each LLM}\label{sec:llm_results}
\begin{table}[!t]
\centering
\caption{Median relative validation and test improvement of \hbbodkl over \texttt{TRIPLE-SH} across ten benchmarks per LLM at different fractions of total LLM calls. IQR in parentheses.}
\label{tab:rel_impro}
\centering
\begin{adjustbox}{width=1.0\columnwidth,center}
\begin{tabular}{lrrrr}
\toprule
 & & \multicolumn{3}{c}{\textbf{Fraction of Total LLM Calls}} \\ \cline{3-5}
 & & 0.25 & 0.50 & 1.00 \\ \midrule
\multirow{2}{*}{\textbf{Claude 3 Haiku}} & Valid & 0.121 (0.145) & 0.059 (0.093) & 0.018 (0.066) \\
& Test & 0.066 (0.105) & 0.027 (0.045) & -0.006 (0.035) \\ \midrule
\multirow{2}{*}{\textbf{LLAMA3 8B Instruct}} & Valid & 0.120 (0.140) & 0.042 (0.086) & 0.001 (0.010) \\
& Test & 0.036 (0.088) & 0.010 (0.036) & 0.000 (0.024) \\ \midrule
\multirow{2}{*}{\textbf{Mistral7B Instruct}} & Valid & 0.068 (0.079) & 0.036 (0.036) & 0.003 (0.044) \\
& Test & 0.039 (0.022) & 0.016 (0.033) & -0.001 (0.047) \\
\bottomrule
\end{tabular}
\end{adjustbox}
\vspace{-1em}
\end{table}

As shown in Section~\ref{sec:main_results}, \texttt{TRIPLE-SH} emerges as the strongest competitor.
To assess whether \hbbodkl's improvements over \texttt{TRIPLE-SH} are consistent across different LLMs, we present in Table~\ref{tab:rel_impro} the median relative improvement over the ten benchmark tasks for each LLM.

The table reveals that \hbbodkl consistently outperforms \texttt{TRIPLE-SH} in terms of both \emph{anytime} validation and test error.
For instance, when using Claude 3 Haiku, the average test error is reduced by a median factor of $0.066$ and $0.027$ at $0.25$ and $0.50$ of the total budget, respectively.
While we further observe positive improvements over \texttt{TRIPLE-SH} on the validation set with a full budget, these gains are less pronounced on the test set.
One possible reason for this discrepancy is that both methods, when given a sufficient budget, successfully identify prompts that have low or optimal validation error.
However, optimal validation error does not necessarily guarantee optimal test error.
We provide further analysis of the generalization gap between validation and test performance that is influenced by the size of the validation set in Appendix~\ref{app:generalization}.

\subsection{Ablation Study}\label{sec:ablation_results}

\begin{figure*}[!t]
    \centering
    \begin{subfigure}{.5\textwidth}
        \includegraphics[width=\linewidth]{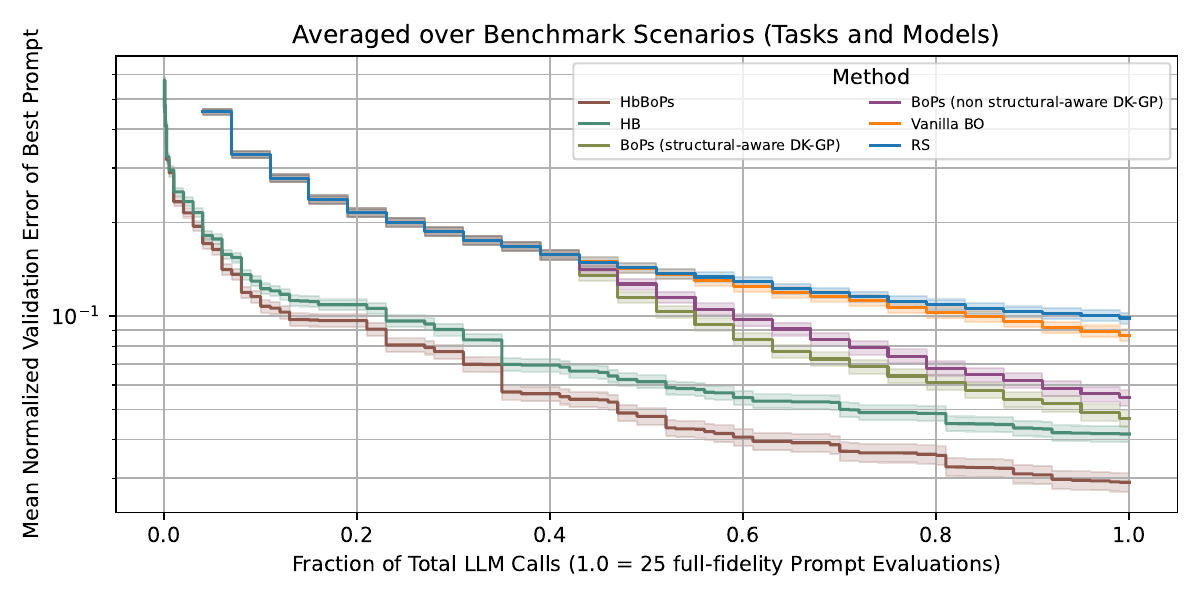}
        \caption{Validation}
        \label{fig:ablation_overall_valid}
        \vspace{-1em}
    \end{subfigure}%
    \begin{subfigure}{.5\textwidth}
        \centering
        \includegraphics[width=\linewidth]{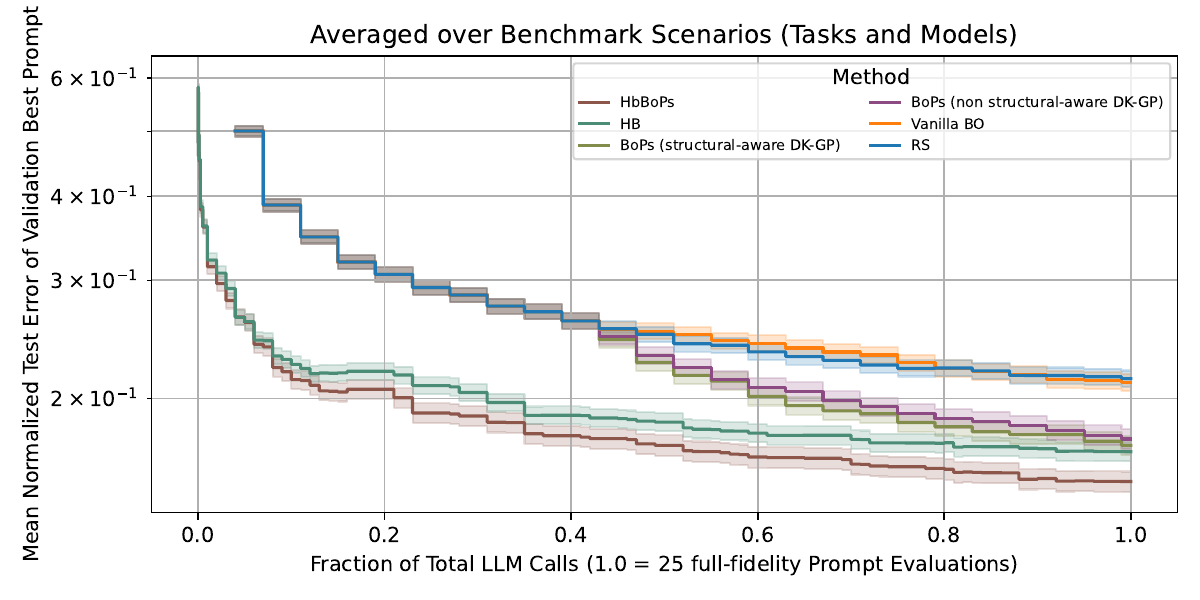}
        \caption{Test}
        \label{fig:ablation_overall_test}
        \vspace{-1em}
    \end{subfigure}%
    \caption{Normalized error (log scale) of the best prompt per \hbbodkl ablation variant, \texttt{RS}, and vanilla \texttt{BO}, averaged over benchmarks. Lower is better. Ribbons represent SE.}
    \label{fig:ablation}
    \vspace{-1em}
\end{figure*}
\begin{table}[!t]
\centering
\caption{Normalized validation and test error of \hbbodkl with different encoders at different fractions of total LLM calls averaged over all 30 benchmarks. SE in parentheses.}
\label{tab:embedding_sensitivity}
\centering
\centering
\begin{adjustbox}{width=1.0\columnwidth,center}
\begin{tabular}{lrrrr}
\toprule
 & & \multicolumn{3}{c}{\textbf{Fraction of Total LLM Calls}} \\ \cline{3-5}
 & & 0.25 & 0.50 & 1.00 \\
\midrule
\multirow{2}{*}{\textbf{BERT}} & Valid & 0.081 (0.004) & 0.048 (0.003) & 0.029 (0.002) \\
& Test & 0.190 (0.006) & 0.170 (0.006) & 0.150 (0.005) \\ \midrule
\multirow{2}{*}{\textbf{MPNet}} & Valid & 0.083 (0.004) & 0.049 (0.003) & 0.031 (0.002) \\
& Test & 0.193 (0.006) & 0.173 (0.006) & 0.158 (0.006) \\ \midrule
\multirow{2}{*}{\textbf{DistillRoBERTa}} & Valid & 0.071 (0.003) & 0.045 (0.002) & 0.026 (0.002) \\
& Test & 0.185 (0.006) & 0.166 (0.006) & 0.150 (0.005) \\ 
\bottomrule
\end{tabular}
\end{adjustbox}
\vspace{-1em}
\end{table}

To better understand the contributions of the individual components of \hbbodkl, we conduct a comprehensive ablation study.
Our ablation focuses on four key aspects: the use of a GP with a deep kernel (DK-GP), the incorporation of a structural-aware DK-GP, the integration of HB for multi-fidelity scheduling, and the final \hbbodkl.
We aim to answer the following main research questions:
(RQ1) Does a structural-aware DK-GP improve over a non-structural-aware DK-GP and a vanilla GP?
(RQ2) Does multi-fidelity scheduling with HB improve over full-fidelity methods?
(RQ3) Does combining our structural-aware DK-GP with HB improve over HB with a random proposal?

Figure~\ref{fig:ablation} presents the average anytime normalized validation and test errors of the best prompt found by systematically removing specific components of our \hbbodkl such that we can quantify their importance and answer above research questions.
We focus on the validation error as shown in Figure~\ref{fig:ablation_overall_valid} to describe results, as improvement of the validation error is a direct consequence of the change of components.

First, we observe that using a DK-GP on prompts embedded as a block of text (\texttt{BoPs (non structural-aware DK-GP)}) in a full-fidelity setting improves over vanilla \texttt{BO} by $11\%$ and $38\%$ at $0.5$ and $1.0$ budget with respect to the average normalized validation error.
This highlights the importance of handling the high-dimensional embedded space properly.
The structural-aware deep kernel (\texttt{BoPs (structural-aware DK-GP)}) further enhances performance by $9\%$ and $13\%$ at $0.5$ and $1.0$ budget, demonstrating the value of directly incorporating structural information into the GP which answers (RQ1).
Note that the structural-aware DK-GP improves over \texttt{HDBO} by $19\%$ and $8\%$ with respect to final average normalized validation and test error.

The integration of HB for multi-fidelity scheduling (\texttt{HB} using a random proposal mechanism) provides orthogonal boosts to both anytime and final performance.
We observe improvements of $47\%$ and $11\%$ at $0.5$ and $1.0$ budget over the full-fidelity \texttt{BoPs (structural-aware DK-GP)} answering (RQ2).

Our complete \hbbodkl further increases performance, achieving a $21\%$ improvement over \texttt{HB} at $0.5$ budget and $31\%$ at $1.0$ budget, answering (RQ3).
Compared to our starting point of vanilla \texttt{BO}, \hbbodkl demonstrates a substantial $66\%$ improvement at $0.5$ budget and $67\%$ at $1.0$ budget.
For additional statistical analyses, we refer to Appendix~\ref{app:additional_results}.

\subsection{Analysis of Varying the Encoder}\label{sec:embedding_sensitivity}
As \hbbodkl relies on embeddings of prompts, we conduct a sensitivity analysis to evaluate the effect of different encoder models on the performance of \hbbodkl.
While our primary results were obtained using BERT's \citep{devlin-etal-2019-bert} [CLS] token embedding, we extend our analysis to include two more encoder models that are MPNet \citep{song2020mpnet} and DistillRoBERTa \citep{liu2019roberta}. 
For each encoder model, we rerun the entire set of benchmarks.

We report in Table~\ref{tab:embedding_sensitivity} the average normalized validation and test error for each encoder when used within \hbbodkl for different fractions of total LLM calls over all 30 benchmark scenarios.
Results show that \hbbodkl maintains consistent validation and test error across all encoder models, indicating robustness to the choice of encoder.
This is expected, as none were specifically fine-tuned for predicting prompt performance.
\hbbodkl's effectiveness stems from its ability to learn a mapping from prompts to performance through the structural-aware DK-GP, provided the embeddings capture meaningful distinctions between prompts.

\section{Related Work}
\label{sec:related_work}

\paragraph{Automating Prompt Engineering.}\label{subsec:prompt_engineering}
Recent work has been concerned with the general topic of automating prompt engineering.
This work can be classified into \emph{prompt optimization}, i.e., automating the creation of prompts \citep{prasad2022grips,sun2022bbtv2,zhou2022large,xu2022gps,diao2022black,chen2023instructzero,lin2023use,fernando2023promptbreeder,pryzant2023automatic,guo2023connecting,pan2023plum,schnabel2024prompts,shen2023reliable,hu2024localized}, and \emph{prompt selection}, i.e., finding the best prompt within a finite candidate set \citep{wu2024prompt,shi2024best,opsahl2024optimizing,do2024automatic}.

Another dimension to categorize the related literature is given by the \emph{white-box} vs. \emph{black-box} setting.
The white-box setting assumes access to the LLM, so that gradient-based methods for prompt optimization or selection are applicable \citep{shin2020autoprompt}.
The black-box setting assumes no access to the LLM which only returns output given input \citep{sun2022black,diao2022black}.

Finally, another differentiation is given by the \emph{static} vs. \emph{dynamic} setting.
The goal of the static setting \citep{wu2024prompt,shi2024best,khattab2023dspy} is to obtain a single prompt offline that in expectation performs well for all instances during test time.
In contrast, the goal of the dynamic setting \citep{zhang2022tempera,do2024automatic,xu2024context} is to select a prompt for each test instance \citep{rubin2021learning} in an online fashion.

\paragraph{Static Black-box Prompt Selection.}\label{subsec:static-black-box-methods}
Our work falls into the category of \emph{static} \emph{black-box} \emph{prompt selection}.
We summarize existing works below.

\texttt{MIPROv2} \citep{opsahl2024optimizing} is \emph{DSPy}'s \citep{khattab2023dspy} state-of-the-art \emph{teleprompter} for joint instruction and few-shot exemplar selection.
It searches over a finite set of candidate prompts by combining instructions with few-shot exemplars (which \emph{DSPy} first constructs automatically).
The method is a variant of BO using a Tree-structured Parzen Estimator (TPE; \citealt{bergstra2011algorithms}) based on the categorical indices of instructions and exemplars ($\text{ID}_i$ and $\text{ID}_e$) that compose a prompt.
A downside is that learning a surrogate model based on indices does not use any semantic information of prompts, which may result in suboptimal predictive performance.
Moreover, \texttt{MIPROv2} does not directly address query-efficiency.
While \emph{DSPy} can be configured to use a smaller random subset of the validation set to evaluate prompts, this risks suboptimal selection due to noisy performance estimates (see also Appendix~\ref{app:generalization}).

\texttt{EASE} proposed by \citet{wu2024prompt} mainly focuses on few-shot exemplar selection.
It uses NeuralUCB (NUCB; \citealt{pmlr-v119-zhou20a}) with embeddings of prompts as blocks of text as features, allowing for sequential evaluation of promising prompts based on the UCB criteria.
\texttt{EASE}'s main contribution is to make the combinatorial problem of selecting examples to build the few-shot exemplar from an \emph{extensive} training set computationally feasible.
It prunes the candidate space using an optimal transport inspired heuristic before applying UCB.
\texttt{EASE} is affected by query-inefficiency since it evaluates prompts on all validation instances (or a random subset, again risking suboptimal selection).


\texttt{TRIPLE} proposed by \citet{shi2024best} is a class of query-efficient algorithms for static black-box prompt selection using a multi-armed bandit approach.
It makes use of Successive Halving \citep{karnin2013almost} or Generalized Successive Elimination \citep{azizi2021fixed} to accelerate prompt evaluation by discarding poor-performing prompts early, reducing the need to evaluate all prompts on all validation instances.
However, \texttt{TRIPLE-SH} is sensitive to the initial evaluation budget and may prematurely discard promising prompts due to noisy performance estimates.
\texttt{TRIPLE-GSE} tries to mitigate this by (non-linear) modeling of expected prompt performance using embeddings of prompts as blocks of text projected to a lower-dimensional space.
While this approach introduces flexibility, the Generalized Successive Elimination algorithm has been formally analyzed only in the generalized linear setting \citep{azizi2021fixed}.
Moreover, both \texttt{TRIPLE-SH} and \texttt{TRIPLE-GSE} begin by evaluating all prompts, whereas our \hbbodkl employs a sample-efficient BO proposal to select candidate prompts for evaluation.

\section{Conclusion}
\label{sec:conclusion}
We introduced \hbbodkl, a method for static black-box prompt selection in which prompts are composed of instructions and few-shot exemplars.
\hbbodkl employs a structural-aware deep kernel Gaussian Process to model the downstream performance of prompts based on separate embeddings of instructions and exemplars.
This enables the identification of promising, unevaluated prompts during the selection process, making \hbbodkl highly sample-efficient.
Furthermore, \hbbodkl integrates Hyperband as a multi-fidelity scheduler that governs the number of validation instances used for prompt evaluation, ensuring query-efficiency.
In extensive experiments, we have demonstrated that \hbbodkl improves upon baselines and state-of-the-art competitors in the limited budget regime while showing strong performance at any stage of the selection process.

While \hbbodkl demonstrates strong performance, some limitations remain.
Our method depends on embeddings from pre-trained encoders.
Although we have demonstrated that \hbbodkl is largely robust to the concrete choice of encoder model, obtaining embeddings induces some minor computational overhead.
Additionally, our analysis focused on prompts composed of instructions and few-shot exemplars.
While these components are highly relevant in practice, prompts may also include additional elements such as output guidance, formatting constraints, or other structural cues, which remain unexplored in this work.
Nevertheless, the use of a deep kernel Gaussian Process as a surrogate model makes our framework, in principle, flexible enough to incorporate such additional prompt parameters by, for example, including categorical variables for formatting styles (e.g., JSON, Markdown).

In our experiments, we evaluated \hbbodkl in a static setting with a fixed set of candidate prompts generated a priori.
This enabled fair comparisons across baselines and methods operating in the static black-box prompt selection setting.
However, we emphasize that \hbbodkl can also be applied in more flexible prompt optimization settings where the candidate pool evolves over time.
For example, it could be integrated with mutation-based prompt generation strategies \citep{fernando2023promptbreeder}, where new prompts are generated iteratively, or with similar end-to-end prompt optimization pipelines \citep{pryzant2023automatic,yang2024opro}.

Future work could extend \hbbodkl to multiple objectives.
Selecting more examples to include in a prompt may enhance performance but also increases response latency.
Balancing the number and composition of examples in a few-shot exemplar introduces a trade-off between performance and efficiency, giving rise to a multi-objective optimization problem.





\section*{Impact Statement}

This paper presents work whose goal is to advance the field of 
Machine Learning. There are many potential societal consequences 
of our work, none which we feel must be specifically highlighted here.


\bibliographystyle{icml2025.bst}
\bibliography{references.bib}

\newpage
\appendix
\onecolumn
\section{On the Latent Space of the Structural-Aware Deep Kernel Gaussian Process}\label{app:dkl}
\begin{figure*}[h]
    \centering
    \begin{subfigure}[b]{0.3\textwidth}
        \centering
        \includegraphics[width=\textwidth]{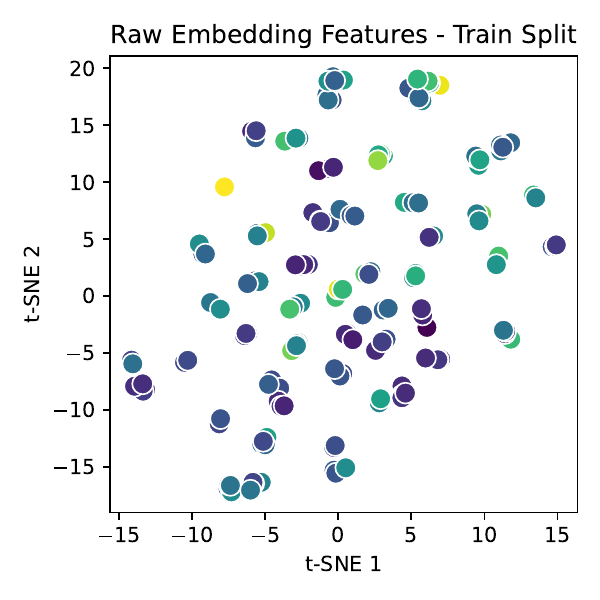}
    \end{subfigure}
    \hfill
    \begin{subfigure}[b]{0.3\textwidth}
        \centering
        \includegraphics[width=\textwidth]{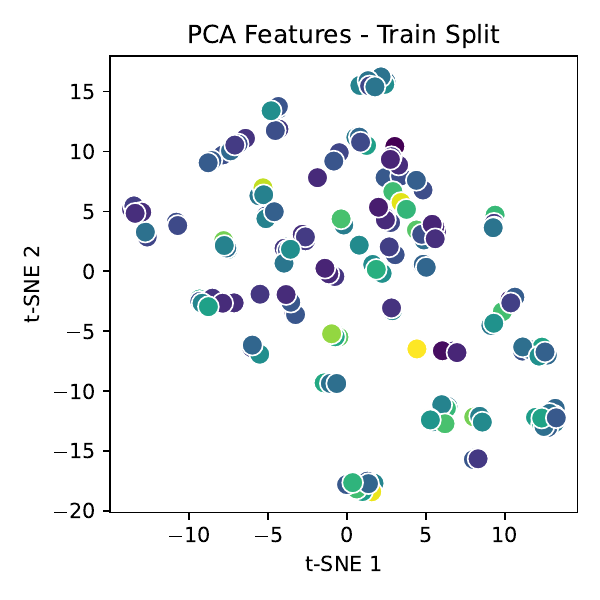}
    \end{subfigure}
    \hfill
    \begin{subfigure}[b]{0.3\textwidth}
        \centering
        \includegraphics[width=\textwidth]{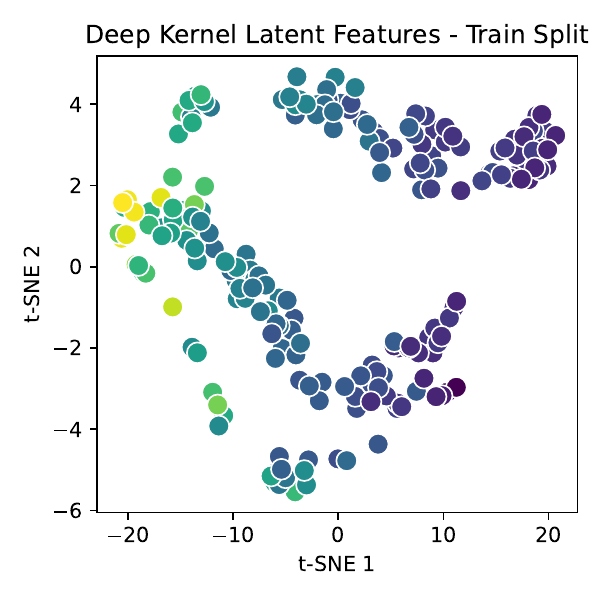}
    \end{subfigure}    
    \begin{subfigure}[b]{0.3\textwidth}
        \centering
        \includegraphics[width=\textwidth]{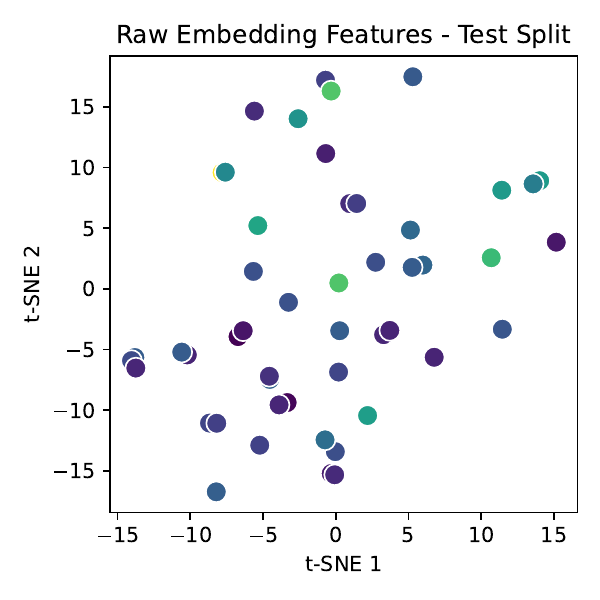}
    \end{subfigure}
    \hfill
    \begin{subfigure}[b]{0.3\textwidth}
        \centering
        \includegraphics[width=\textwidth]{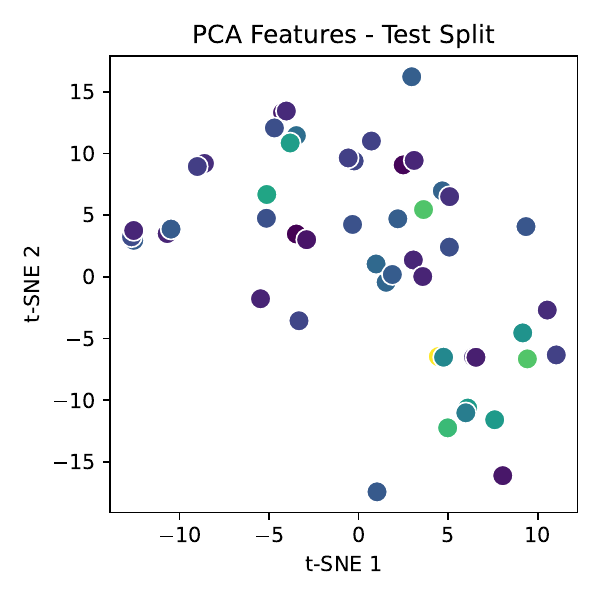}
    \end{subfigure}
    \hfill
    \begin{subfigure}[b]{0.3\textwidth}
        \centering
        \includegraphics[width=\textwidth]{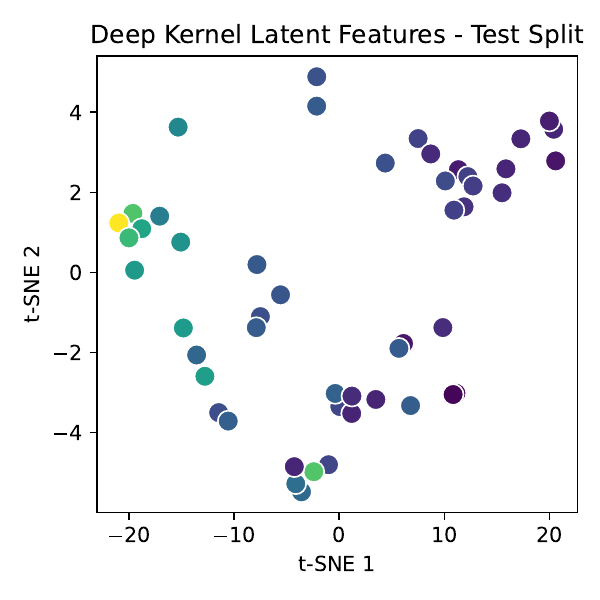}
    \end{subfigure}
    \caption{Visualization of the $768$ dimensional BERT [CLS] token embeddings of prompts via a two component t-SNE. Left: Raw, unprocessed features. Middle: Features of a $10$ component PCA solution. Right: Latent features ($10$ dimensional) from the feature extractor of our structural-aware DK-GP. Top row: Train split. Bottom row: Test split. Color indicates the performance of prompts for LLAMA3 8B Instruct on \emph{GSM8K}.}
    \label{fig:feature_comparison}
\end{figure*}

As described in Section~\ref{sec:gp}, unsupervised dimensionality reduction techniques such as PCA or random projections will not result in a lower-dimensional latent representation of prompt embeddings that is aligned with the downstream performance of prompts.
To illustrate this, we perform the following experiment:
We collect the validation error of 250 prompts on \emph{GSM8K} (according to the splits described in Table~\ref{tab:tasks}) using LLAMA3 8B Instruct as LLM.
We embed each prompt using the [CLS] token embedding of BERT ($d = 768$).
We then split the prompts $p_{1}, \ldots, p_{250}$ and their corresponding validation errors $v_{1}, \ldots, v_{250}$ in a train (80\%) and test set (20\%).
Using the train split, we perform a PCA and retain $10$ principal components as features.
Moreover, we train our structural-aware DK-GP introduced in Section~\ref{sec:structural_aware_dkl} on the training split and extract the $10$ latent features from the output of the feature extractor $\phi_{\left( \phi_{enc(i)}, \phi_{enc(e)} \right)}$.
We visualize the raw $768$ dimensional embedding features of prompts, the $10$ dimensional PCA features and the $10$ dimensional deep kernel features for the training split using a two component t-SNE \citep{van2008visualizing} in the top row of Figure~\ref{fig:feature_comparison}.
The $x$- and $y$-axis represent the two t-SNE components, whereas color indicates the validation error of prompts (lighter color indicates better performance).
We can see that for both the raw embedding features and the PCA features, it is difficult to visually discern any meaningful clusters or structure how closeness in feature space relates to closeness in performance space.
For the deep kernel features, however, we can see that the feature space is well aligned with the performance space (well-performing prompts being closer together with a continuous transition into poorer performing prompts) - of course, this is on the training split on which the GP has been trained on, and therefore these results are not surprising.
However, when looking at the test split (that was neither used to perform the PCA nor to train the GP) in the bottom row of Figure~\ref{fig:feature_comparison}, we can see that similar conclusions as for the train split hold:
The latent representation the feature extractor of the deep kernel has learned during training does generalize to the test split, and it has effectively learned a low-dimensional embedding of prompts aligned with the downstream task.

\section{On the Generalization from Validation to Test}\label{app:generalization}
When performing black-box prompt selection, we evaluate prompts on a validation set and iteratively improve over the current best prompt (the incumbent), trying to identify a better one.
While progress on the validation set is expected (i.e., if we perform full-fidelity evaluations using the same validation instances for each prompt, the validation error of the incumbent will be monotonically decreasing as optimization progresses), it must not necessarily be the case that we also improve performance on a held-out test set of instances, i.e., the prompt identified as being validation optimal might not necessarily be optimal on the test set.

In this section, we provide additional insights regarding generalization gaps from validation to test performance.
Recall that existing methods for black-box prompt selection (e.g., \texttt{EASE} and \texttt{MIPROv2}) are by design not query-efficient but evaluate all prompts on all validation instances or a random subset (e.g., \citealt{wu2024prompt} used sub-sampled validation sets with as few as 20 instances).
We now empirically demonstrate that using small random subsets during optimization is not a sensible choice, because this will result in increased variance of the estimate of the validation error, which prevents us from making correct decisions on the validation set.
This can result in generalization issues when moving from the validation set to the test set.

\begin{figure}[h]
    \centering
    \begin{subfigure}[b]{0.19\linewidth}
        \includegraphics[width=\linewidth]{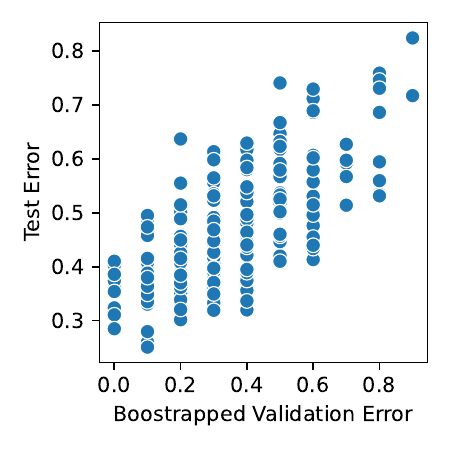}
        \caption{$k = 10$.}
    \end{subfigure}
    \begin{subfigure}[b]{0.19\linewidth}
        \includegraphics[width=\linewidth]{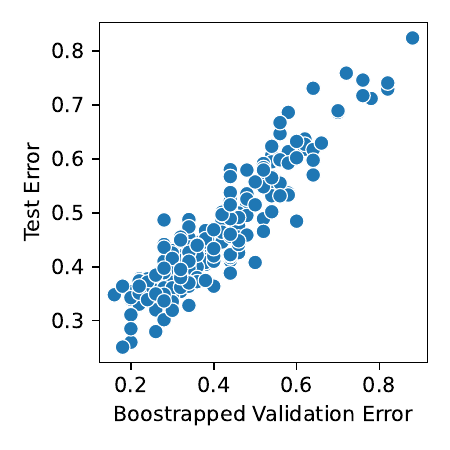}
        \caption{$k = 50$.}
    \end{subfigure}
    \begin{subfigure}[b]{0.19\linewidth}
        \includegraphics[width=\linewidth]{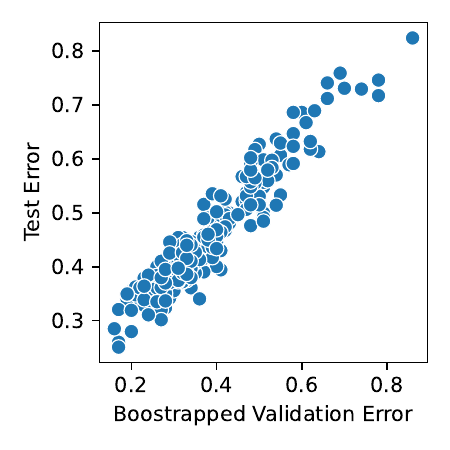}
        \caption{$k = 100$.}
    \end{subfigure}
    \begin{subfigure}[b]{0.19\linewidth}
        \includegraphics[width=\linewidth]{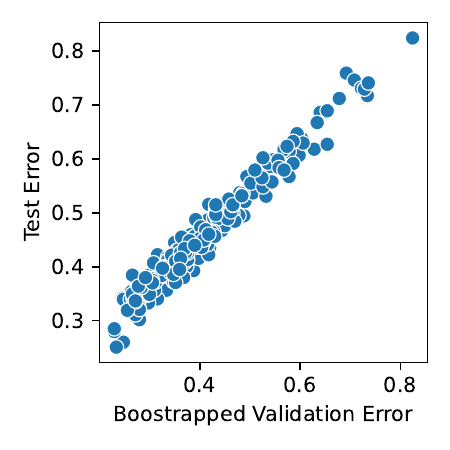}
        \caption{$k = 500$.}
    \end{subfigure}
    \begin{subfigure}[b]{0.19\linewidth}
        \includegraphics[width=\linewidth]{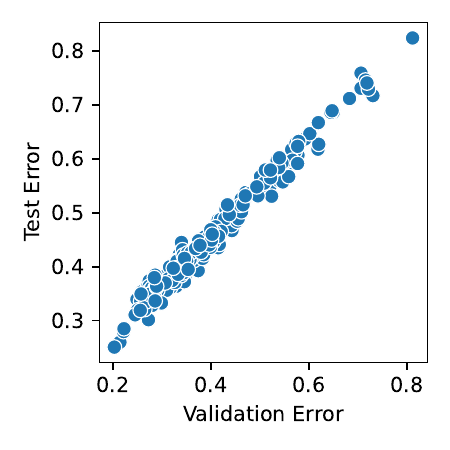}
        \caption{Full validation set.}
    \end{subfigure}
    \caption{Scatter plots of the validation and test errors of 250 prompts evaluated with LLAMA3 8B Instruct on \emph{GSM8K} using differently sized ($k = 10, 50, 100, 500)$ bootstrap samples of validation instances (a) to (d) or the full validation set (e).}
    \label{fig:gsm8k_llama3_valid_test}
\end{figure}

We perform the following experiment:
We collect the validation and test error (according to the splits described in Table~\ref{tab:tasks}) of 250 prompts on the \emph{GSM8K} task using LLAMA3 8B Instruct as LLM.
We vary the number of validation instances used to evaluate the performance of prompts via bootstrapping, using $k = 10, 50, 100, 500$ instead of the original $n_{\mathrm{valid}} = 1319$ validation instances.
Note that to compute the test error, we always use the full test set.
In Figure~\ref{fig:gsm8k_llama3_valid_test}, we provide scatter plots of the validation and test errors of the prompts with mean validation errors obtained via bootstrapping using $k = 10, 50, 100, 500$ validation instances vs. validation errors obtained on the full validation set of $1319$ instances.
We perform $1000$ bootstrap replicates.

We can observe that if we use too few validation instances ($k = 10$ but also $k = 50$ and $k = 100$ to some extent) even if we select the validation optimal prompt, its test error can be far from optimal, because noise dominates the estimate of the validation error.

\begin{figure}[t]
    \centering
    \includegraphics[width=0.5\linewidth]{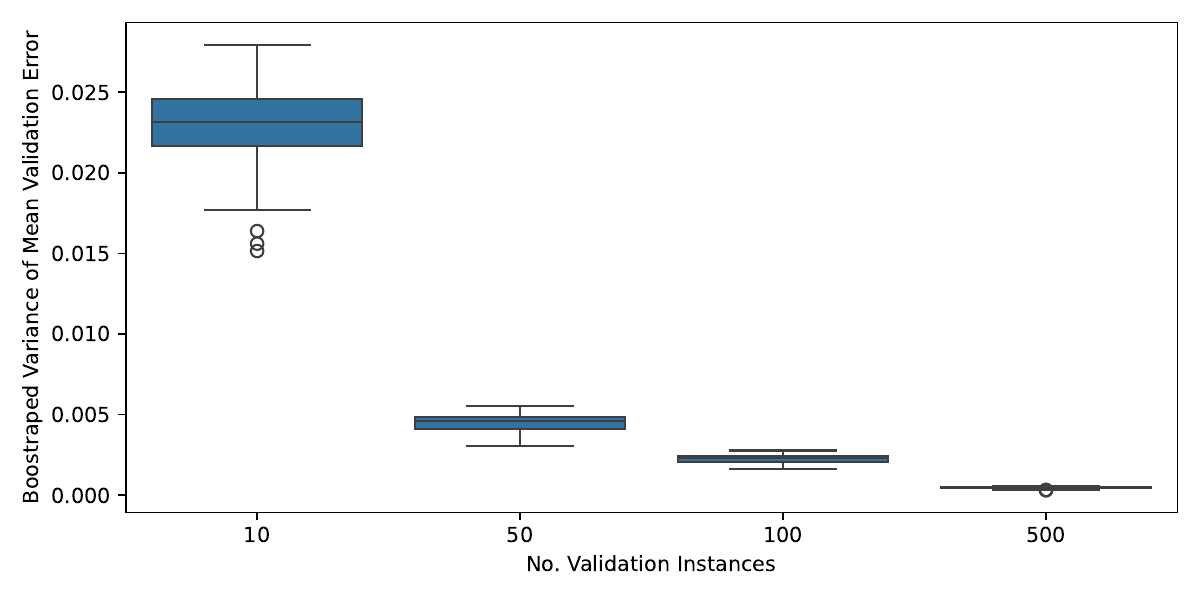}
    \caption{Box plots of the bootstrapped variance estimates of the mean validation error of 250 prompts evaluated with LLAMA3 8B Instruct on \emph{GSM8K} varying the number of validation instances used to estimate the mean validation error.}
    \label{fig:gsm8k_llama3_valid_variance_bootstrap}
\end{figure}

In Figure~\ref{fig:gsm8k_llama3_valid_variance_bootstrap}, we provide box plots of the bootstrapped variance estimates of the mean validation error over prompts when using smaller validation sets.
As expected, the variance of the mean validation error can be substantial when using few validation instances.
Note that the bootstrap results align with theoretical expectations under the assumption that the point-wise loss (based on exact match) is a binary random variable following a Bernoulli distribution with success probability $p$ (corresponding to a loss of $0$).
In this case, the average validation error over $n_{\mathrm{valid}}$ instances follows a Binomial distribution, and the variance of the estimated validation error is given by $\nicefrac{p(1-p)}{n_{\mathrm{valid}}}$.
For example, if a prompt has a true ``success probability'' of $p=0.5$, then using $n_{\mathrm{valid}} = 10$ validation instances yields an expected variance of $\nicefrac{0.5^2}{10} = 0.025$.

This has serious practical implications, depending on the variation in the performance of prompts on a downstream task.
If the true performance of many prompts is similar, we cannot tell them apart based on their estimated validation error, as noise dominates the signal when using too few validation instances.
Moreover, for benchmarking methods for black-box prompt selection this is highly relevant as when using too few validation instances, we cannot determine whether a generalization gap from the validation set to the test set results solely from noisy performance estimates or from internal method mechanisms (such as the optimal transport inspired heuristic employed by \texttt{EASE} to only consider examples that are similar to the validation set) which may further result in overfitting to the validation set.

The empirical results we have presented here further provide justification for using a multi-fidelity scheduler over the validation instances during prompt selection, such as Hyperband.
Poor-performing prompts can be differentiated using few validation instances, however, well-performing prompts need to be evaluated on larger sets so that one can effectively tell their validation errors apart.
\section{Multi-Fidelity over Validation Instances}\label{app:mf}
In this section, we discuss how one can define a multi-fidelity schedule for prompt selection, in which the fidelity parameter is the number of validation instances.
Let $\mathcal{D}_{\mathrm{valid}} = \{(x_i, y_i)\}_{i=1}^{n_{\mathrm{valid}}}$ denote the validation set containing $n_{\mathrm{valid}}$ input-output instances in natural language on which the LLM $h_{p}$ configured to use a given prompt $p \in \mathcal{P}$ is evaluated.
Recall our goal of identifying the optimal prompt: $\argmin_{p \in \mathcal{P}} \mathbb{E}_{(x, y) \sim \mathbb{P}_{xy}} \left[ l(y, h_{p}(x)) \right]$.
Here, the expectation is taken over all input output instances from a data-generating distribution $\mathbb{P}_{xy}$ and $l$ is the pointwise loss function used to compare the LLM's output $h_{p}(x)$ to the ground truth $y$.
We want to identify the best performing prompt while minimizing the number of LLM calls for evaluation, given the significant costs in both time, but especially query expenses associated with LLM black-box APIs.

In practice, we can only approximate this expectation by the full-fidelity evaluation on the whole validation set given by $\frac{1}{n_{\mathrm{valid}}} \sum_{i=1}^{n_{\mathrm{valid}}} l(y_i, h_{p}(x_i))$.

The idea of multi-fidelity techniques is to speed up and reduce the cost of the evaluation of prompts by using fewer validation instances during evaluation.
Let $\mathcal{V} = \{1, \ldots, n_{\mathrm{valid}}\}$ denote the index set corresponding to the indices of validation instances.
A simple way to reduce cost of evaluation is to use a random subset, $\mathcal{S} \subset \mathcal{V}$ of validation instances: $\frac{1}{|\mathcal{S}|} \sum_{i \in \mathcal{S}} l(y_i, h_{p}(x_i))$.
This was for example done in the evaluation protocol of the benchmark study in \citet{wu2024prompt}.
However, it comes with the following downsides:
(1) It is a priori not clear how many instances are needed to obtain an accurate estimate of the validation error of prompts as this depends on the concrete LLM, downstream task, and variation of the true performance over prompts.
(2) Using a fixed (sub-)sample is inefficient because all prompts are evaluated on the same number of validation instances.
However, poor-performing prompts can be identified using few validation instances, whereas well-performing prompts should be evaluated on many validation instances to differentiate between them and not risk suboptimal selection.

\citet{shi2024best} made the connection between multi-fidelity prompt selection and best arm identification from the multi-armed bandit literature.
In this setting, ``pulling an arm'' refers to evaluating a prompt on a validation instance.
The goal is to identify the best arm within a limited budget of evaluations.
A well-known algorithm from the bandit literature is given by Successive Halving (SH, \citealt{karnin2013almost,jamieson2016non}).
In our setting of prompt selection, the idea of SH is to efficiently identify the best performing prompt (arm) under a given budget constraint of total LLM calls (pulls).
Given a total budget $B$ of LLM calls and $n \coloneqq |\mathcal{P}|$ prompts, SH starts by allocating a budget of $b \coloneqq B / (n \log_{2}(n))$ LLM calls to each prompt.
After having evaluated each prompt on $b$ instances, the lower half of bad performing prompts are discarded and the process repeats, doubling the number of LLM calls for the remaining prompts in the next stage.
This process in general repeats until a single prompt remains.
SH can be employed in both a stochastic and non-stochastic setting.
The stochastic setting \citep{karnin2013almost} is formally characterized by the following:
(1) Losses are i.i.d. samples from a probability distribution.
(2) Each arm has a fixed expected loss $\mu_{p}$.
(3) The goal is to identify the arm with the lowest expected loss.

\citet{jamieson2016non} introduced SH in the non-stochastic setting when applying it to the problem of hyperparameter optimization, where the budget can, for example, be the size of the training set used to train an algorithm or the number of epochs used to train a neural network.
This non-stochastic setting formally is characterized by the following:
(1) Losses are real numbers chosen by an oblivious adversary.
(2) Each arm has a limit $\nu_{p}$ of its loss sequence as the number of evaluations go to infinity.
(3) The goal is to identify the arm with the lowest loss limit.

Within the non-stochastic setting, \citet{li2018hyperband} introduced Hyperband (HB) although it is in general also applicable to the stochastic setting.
Recall that SH requires the overall budget $B$ and the number of prompts $n$ as input parameters.
These determine the starting budget $b$ (number of validation instances) prompts are evaluated on.
However, for a given overall budget $B$ it is a priori not clear whether one should evaluate many prompts using on average few validation instances (resulting in better exploration of the search space at the risk of noisy validation error estimates) or whether one should consider a smaller number of prompts using on average more validation instances (focusing on fewer prompts but obtaining more accurate validation error estimates).

Our proposal to use HB as a multi-fidelity scheduler for prompt selection improves over SH by addressing this trade-off between the number of prompts to explore and the amount of resources to allocate to each configuration.
It does so by running multiple SH brackets, each with a different initial number of prompts and initial per-prompt budget.
This allows HB to efficiently explore the search space, quickly discarding poor prompts while allocating more resources to promising ones, therefore hedging against a poor choice of the number of prompts and per-prompt budget.

We present pseudocode for HB adapted to the problem of black-box prompt selection in Algorithm~\ref{alg:hyperband_modified_sequential} (as explained in Section~\ref{sec:hbbodkl}, \hbbodkl replaces the random proposal mechanism of HB by a BO proposal) and in Table~\ref{tab:hyperband_example}, we present an exemplary schedule describing how the number of prompts used in each stage of each bracket of the algorithm relates to the number of validation instances used to evaluate the prompts.
Regarding the output of the algorithm, we note that there is a critical design decision:
The vanilla HB \citep{li2018hyperband} algorithm would output the configuration with the smallest validation error observed so far.
However, in the context of prompt selection, validation errors are determined based on different numbers of validation instances varying between stages of brackets ($\{s_{\max}, s_{\max} - 1, \ldots, 0\}$).
To allow for a robust selection of the optimal prompt, we always return the prompt with the lowest validation error among all prompts that were evaluated on the full validation set\footnote{Or evaluated on the largest subset used so far when assessing the anytime performance of the algorithm.}.
This is crucial for robust performance of HB in the context of prompt selection, which we demonstrate in Appendix~\ref{app:additional_results_hb_ablation}.

\begin{table}[h]
\centering
\caption{Exemplary HB schedule for black-box prompt selection assuming a minimum budget of $b_{\min} = 10$ validation instances, a maximum number of $n_{\mathrm{valid}} = 80$ validation instances being available in total, and a halving parameter of $\eta = 2.0$.}
\label{tab:hyperband_example}
\begin{tabular}{cccc}
\hline
\textbf{Bracket} ($s$) & \textbf{Stage} ($i$) & \textbf{\#Instances} ($b$) & \textbf{\#Prompts} ($n$) \\
\hline
3 & 0 & 10 & 8 \\
3 & 1 & 20 & 4 \\
3 & 2 & 40 & 2 \\
3 & 3 & 80 & 1 \\ \midrule
2 & 0 & 20 & 6 \\
2 & 1 & 40 & 3 \\
2 & 2 & 80 & 1 \\ \midrule
1 & 0 & 40 & 4 \\
1 & 1 & 80 & 2 \\ \midrule
0 & 0 & 80 & 4 \\
\hline
\end{tabular}
\end{table}

Our HB adapted to black-box prompt selection has three inputs: $n_{\mathrm{valid}}$ the total number of validation instances available (depending on the task), $b_{\mathrm{min}}$ a lower bound on the number of validation instances used for prompt evaluation, and $\eta$ the halving parameter of the SH subroutine.
HB makes use of three functions:\\
(1) $\text{get\_prompt}()$ returns a candidate prompt from the search space $\mathcal{P}$. In vanilla HB, we would sample uniformly at random. In our \hbbodkl we obtain the next candidate prompt via a BO proposal (Section~\ref{sec:hbbodkl}).\\
(2) $\text{get\_validation\_error}(p, b_i)$ evaluates a prompt $p$ using $b_i$ validation instances.
We further discuss this below.\\
(3) $\text{top\_k}(P, V, \lfloor n_i/\eta \rfloor)$ reduces the $n_i$ prompts in the active set $P$ by only keeping the $\lfloor n_i/\eta \rfloor$ best performing ones.

As mentioned above, there is another critical detail when adapting HB to the problem of prompt selection, which is concerned with the selection of validation instances for a given stage of a bracket but also when moving from a given stage to the next stage within a bracket.
In principle, the validation instances in $\text{get\_validation\_error}(p, b_i)$ could be different for each prompt (sampled uniformly at random from the set of validation instances).
However, to allow for a fairer comparison of the performance of prompts, we decided to use the same, fixed subset of validation instances for each prompt evaluated at a given stage of a bracket.
This ensures that we perform a paired comparison of the performance of prompts when discarding the worst performing half.
We ablate this design decision in Appendix~\ref{app:additional_results_hb_ablation}.

Moreover, when moving from one stage to another stage, there are two possibilities how to construct the subset of validation instances used in the next stage:
(1) Simply draw a sample.
(2) Keep the already used validation instances of the previous stage and only sample the remaining number of additional validation instances needed to fill the current stage from the remaining yet not used validation instances.
The second option is highly desirable to reduce the total number of LLM calls if we cache the evaluation of prompts.
The validity of this modeling choice naturally depends on the degree of stochasticity in the LLM's output for a given prompt-instance pair, which in turn is influenced by the sampling temperature.
However, if we assume reasonably deterministic outcomes, we can reduce the total number of LLM calls used in HB drastically (i.e., roughly by a factor of $\eta$),
In \hbbodkl we cache the output for a given prompt and validation instance and reuse the already used validation instances of the previous stage and only sample the remaining needed validation instances.
We ablate this design decision in Appendix~\ref{app:additional_results_hb_ablation}.

We want to note that \hbbodkl employs a fully sequential HB schedule, i.e., prompts are proposed sequentially, and we evaluate brackets and their stages in their given order (e.g., as described in Table~\ref{tab:hyperband_example}).
While plenty possibilities exist to parallelize SH or HB in the context of hyperparameter optimization where evaluating a configuration involves a training step, we argue that in the context of prompt selection, there is little gain made by performing a batch proposal of prompts and evaluating batch-parallel (as in vanilla HB for hyperparameter optimization) or using asynchronous multi-fidelity schedulers \citep{li2020system}.
This is because parallelization can be directly performed on the lowest level of evaluating a prompt on the validation set, i.e., in $\text{get\_validation\_error}(p, b_i)$ by parallelizing LLM calls when evaluating a prompt $p$ on the $b_i$ validation instances.
As a final comment on overhead, note that it may initially seem that training the structural-aware DK-GP in each iteration of a BO proposal within \hbbodkl leads to computational costs that increase proportionally as optimization progresses.
However, the GP is trained only on the subset of the highest fidelity design data for which enough observations are available.
This approach not only ensures that the design data used to train the GP has accurately estimated validation errors but also keeps the subset size manageable, mitigating scaling issues of the GP even when \hbbodkl is executed repeatedly.
\section{Details on the Experimental Setup}\label{app:experiments}

\subsection{Tasks}
Table~\ref{tab:tasks} reports characteristics on the benchmark tasks used in our experiments.
For \emph{AI2 ARC}, we use the official train, validation and test splits from AI2’s Reasoning Challenge.
However, during inspection we noticed that for some reason, AI2's Reasoning Challenge includes a few instances with choices named ``1'', ``2'', ``3'', ``4'' instead of ``A'', ``B'', ``C'', ``D'' and a few instances with five choices instead of four which we excluded from the splits for consistency.
\emph{GSM8K} officially only contains a train and test split.
We sampled $1319$ instances from the train split uniformly at random to create a validation set of comparable size to the test set.
For all other tasks from the BBII subset of the BIG-bench and instruction induction benchmarks, we use the splits as proposed by \citet{wu2024prompt}.
Unlike \citet{wu2024prompt}, who used only 20 validation instances by further sub-sampling the validation splits, we retain larger validation splits to reduce noise and improve the reliability of performance estimates.
For each task, the training split was used to generate instructions for the prompts via APE's \citep{zhou2022large} forward mode and to select instances for few-shot exemplars.
Instructions are generated using APE’s forward mode, where Claude 3 Sonnet \citep{TheC3}, configured with a temperature of $1.0$ and default settings otherwise (as in the main paper for Claude 3 Haiku), produces $100$ candidate instructions from ten input-output examples per task; these are then embedded using BERT's [CLS] token representation, and five representative instructions are selected via 5-medoid clustering.
The validation split was used during optimization and the test split was used to assess unbiased performance.
We perform standard sanitization of LLM outputs to be able to employ the loss function in Equation~\eqref{eq:em_loss} based on the exact match as a scoring function.
For \emph{GSM8K}, we determine the prediction for the exact match loss function by selecting the last number contained in the LLM's output.
This approach aligns with the Chain-of-Thought prompting and the typical output behavior of LLMs for this task.

\begin{table}[h]
\centering
\caption{Characteristics of tasks used in the experiments.}
\label{tab:tasks}
\begin{tabular}{lrrrr}
\toprule
\textbf{Task} & \textbf{Setting} & $n_{\mathrm{train}}$ & $n_{\mathrm{valid}}$ & $n_{\mathrm{test}}$ \\
\midrule
AI2 ARC                 & multiple choice question answering & $1094$ & $291$ & $1144$ \\
GSM8K                   & grade school math questions & $6154$ & $1319$ & $1319$ \\
antonyms                & find antonym of word & $2073$ & $519$ & $100$ \\
larger animal           & select larger of two animals & $2422$ & $606$ & $100$ \\
negation                & negate a sentence & $723$ & $181$ & $100$ \\
object counting         & count number of objects & $560$ & $140$ & $100$ \\
orthography starts with & output all words starting with a given letter & $2400$ & $600$ & $100$ \\
second word letter      & output the second letter of a word & $2644$ & $662$ & $100$ \\
sentiment               & sentiment analysis of movie rating & $933$ & $234$ & $100$ \\
word unscrambling       & build a word from scrambled letters & $5627$ & $1407$ & $100$ \\
\bottomrule
\end{tabular}
\end{table}

\subsection{LLMs}
We use Claude 3 Haiku \citep{TheC3}, LLAMA3 8B Instruct \citep{dubey2024llama}, and Mistral 7B Instruct \citep{jiang2023mistral} with default hyperparameters (Claude 3 Haiku: max tokens $= 200$, temperature $= 0.5$, top p $= 1.0$, top $k = 250$; LLAMA3 8B Instruct: max tokens $= 512$, temperature $= 0.5$, top $p = 0.9$; Mistral 7B Instruct: max tokens $= 512$, temperature $= 0.5$, top p $= 0.9$, top $k = 50$).
For \emph{GSM8K} we increase max tokens for all LLMs to $1024$.

\subsection{Methods}
We run all methods as described in Section~\ref{sec:experimental_setup}.
All full-fidelity BO methods are implemented within BoTorch \citep{balandat2020botorch}.
We include \texttt{HDBO} \citep{hvarfner2024vanilla} to have a simple yet well-performing ``high-dimensional'' BO baseline.
\citet{hvarfner2024vanilla} recently challenged the general belief that vanilla BO does not perform well for high-dimensional functions by training a GP via MAP with priors over kernel and likelihood parameters adjusted to reflect the dimensionality of the problem which resulted in strong BO performance on high-dimensional functions.
We run \texttt{EASE} as implemented in the official code base\footnote{\url{https://github.com/ZhaoxuanWu/EASE-Prompt-Optimization/blob/e3514de58bd682ebc5ea46fe890481f2b92e5589/experiments/LlamaForMLPRegression.py}}, \texttt{MIPROv2} as implemented in \texttt{DSPy}\footnote{\url{https://github.com/stanfordnlp/dspy/blob/425b6f07d5cf0530f5a5566ad4f247b15aecb522/dspy/teleprompt/mipro_optimizer_v2.py}} and \texttt{TRIPLE-SH} and \texttt{TRIPLE-GSE} as implemented in the official code base\footnote{\url{ https://github.com/ShenGroup/TRIPLE/blob/06264a97b4dd766c9a88afc24058627fac0f223d/src/bandit/contextual/gse.py}}.
Regarding \texttt{TRIPLE-GSE} we noticed that \citet{shi2024best} did not provide detailed descriptions of the model used to predict expected prompt performance (in the main paper, they state that one can use a linear model or MLP).
In their implementation, however, they use an ensemble of a Bayesian ridge regression model, a gradient boosting regression model, and an MLP where weights for the ensemble are determined based on each model's $R^2$ performance on a separate validation set.
We did not change this modeling approach when running \texttt{TRIPLE-GSE}.
We implement our \hbbodkl in GPyTorch \citep{gardner2018gpytorch} and run it as described in Section~\ref{sec:experimental_setup}.
Moreover, to hedge against poor model-based proposals, we perform random interleaving as described in \citet{falkner2018bohb} for each proposal with a probability of $\rho = 0.1$.
Since a single SH or HB schedule may require less budget than the total pre-defined LLM call budget per task, we repeatedly run all methods that use SH or HB as multi-fidelity schedulers until their total LLM call usage reaches this pre-defined budget.
\section{Additional Results}\label{app:additional_results}

\subsection{Main Results}\label{app:additional_results_main}
Here, we provide additional analyses of the main results reported in Section~\ref{sec:main_results}.
To test whether \hbbodkl outperforms all other methods with respect to validation and test error for different fractions of budget, we conduct a linear mixed effects model analysis.
We model the unaggregated performance as a function involving random intercepts for each benchmark scenario (benchmark task and LLM combination).
This approach is sensible as each method has been run repeatedly on the same benchmark scenario with different random seeds that affect, for example, initial designs.
To test the global hypothesis that there is an effect of the method on performance, we test an intercept model against a model including an effect of the factor method.
If we reject the null hypothesis, we proceed with a Tukey post hoc test (corrected for multiple testing) to test each method against \hbbodkl.
We test at the conservative $\alpha = 0.01$ level.

For the validation error at a fraction of $1.00$, we reject the global null hypothesis of no effect of methods ($\chi^2(8) = 672.95, p < 1\mathrm{e}{-4}$).
The pairwise results are:
{\tiny
\begin{itemize}
    \item \texttt{RS} vs. \hbbodkl, $z = 18.52, p < 1\mathrm{e}{-4}$
    \item vanilla \texttt{BO} vs. \hbbodkl, $z = 15.40, p < 1\mathrm{e}{-4}$ 
    \item \texttt{HDBO} vs. \hbbodkl, $z = 7.65, p < 1\mathrm{e}{-4}$
    \item \texttt{BOPCA} vs. \hbbodkl, $z = 9.95, p < 1\mathrm{e}{-4}$ 
    \item \texttt{EASE} vs. \hbbodkl, $z = 17.29, p < 1\mathrm{e}{-4}$ 
    \item \texttt{MIPROv2} vs. \hbbodkl, $z = 13.18, p < 1\mathrm{e}{-4}$ 
    \item \texttt{TRIPLE-SH} vs. \hbbodkl, $z = 1.87, p = 0.236$ 
    \item \texttt{TRIPLE-GSE} vs. \hbbodkl, $z = 6.26, p = 1\mathrm{e}{-4}$ 
\end{itemize}
}
We conclude that \hbbodkl outperforms all methods significantly with respect to final performance, except for \texttt{TRIPLE-SH}.
While the effect is positive, i.e., \hbbodkl improves over \texttt{TRIPLE-SH}, it  is not strong enough to be considered statistically significant at the $\alpha = 0.01$ level.
For brevity, we do not include results for fractions of $0.25$ or $0.50$ here, where we observed similar results but \hbbodkl more strongly outperforming the other methods.

\begin{table}[h]
\centering
\caption{Normalized validation error of each method on each benchmark for Claude 3 Haiku. Averaged over repetitions. Standard errors in parentheses. For visual analysis, we highlight all methods that have a mean error that is less or equal to the mean error of the best method plus two times its standard error.}
\label{tab:raw_valid_haiku}
\centering
\begin{adjustbox}{width=1.00\linewidth,center}
\begin{tabular}{lrrrrrrrrrr}
\toprule
& \multicolumn{10}{c}{\textbf{Benchmark}}\\ \cline{2-11}
 & AI2 ARC & GSM8K & antonyms & larger animal & negation & object counting & orthography starts with & second word letter & sentiment & word unscrambling \\
\midrule
RS & 0.053 (0.007) & 0.087 (0.010) & 0.018 (0.002) & 0.011 (0.002) & \textbf{0.000 (0.000)} & 0.093 (0.011) & 0.033 (0.005) & 0.076 (0.011) & 0.021 (0.003) & 0.085 (0.007) \\
vanilla BO & \textbf{0.046 (0.007)} & 0.084 (0.010) & 0.020 (0.002) & 0.012 (0.002) & \textbf{0.000 (0.000)} & 0.075 (0.010) & 0.033 (0.005) & 0.097 (0.011) & 0.021 (0.004) & 0.084 (0.009) \\
HDBO & \textbf{0.039 (0.007)} & 0.050 (0.007) & \textbf{0.014 (0.002)} & \textbf{0.004 (0.001)} & \textbf{0.000 (0.000)} & \textbf{0.040 (0.009)} & 0.028 (0.004) & 0.029 (0.008) & \textbf{0.014 (0.003)} & 0.039 (0.008) \\
BOPCA & \textbf{0.047 (0.008)} & \textbf{0.041 (0.007)} & \textbf{0.014 (0.002)} & \textbf{0.005 (0.001)} & \textbf{0.000 (0.000)} & 0.062 (0.011) & 0.035 (0.005) & 0.044 (0.010) & \textbf{0.015 (0.003)} & 0.081 (0.009) \\
EASE & 0.080 (0.010) & 0.063 (0.003) & 0.024 (0.002) & 0.017 (0.002) & \textbf{0.000 (0.000)} & 0.099 (0.010) & 0.039 (0.005) & 0.070 (0.010) & 0.027 (0.003) & 0.103 (0.007) \\
MIPROv2 & \textbf{0.048 (0.008)} & 0.052 (0.004) & 0.016 (0.001) & 0.011 (0.002) & \textbf{0.000 (0.000)} & \textbf{0.050 (0.009)} & 0.019 (0.003) & 0.043 (0.009) & 0.025 (0.003) & 0.059 (0.007) \\
TRIPLE-SH & 0.109 (0.014) & \textbf{0.043 (0.008)} & 0.029 (0.003) & \textbf{0.005 (0.002)} & 0.003 (0.002) & \textbf{0.043 (0.010)} & 0.035 (0.006) & \textbf{0.006 (0.002)} & 0.051 (0.007) & \textbf{0.025 (0.006)} \\
TRIPLE-GSE & 0.178 (0.014) & 0.072 (0.010) & 0.031 (0.005) & 0.021 (0.003) & 0.000 (0.000) & 0.087 (0.014) & 0.049 (0.008) & \textbf{0.007 (0.002)} & 0.060 (0.009) & \textbf{0.016 (0.006)} \\
\hbbodkl & \textbf{0.040 (0.008)} & \textbf{0.035 (0.005)} & \textbf{0.011 (0.002)} & \textbf{0.005 (0.001)} & \textbf{0.000 (0.000)} & \textbf{0.055 (0.011)} & \textbf{0.013 (0.002)} & \textbf{0.008 (0.002)} & 0.023 (0.003) & \textbf{0.026 (0.005)} \\
\bottomrule
\end{tabular}
\end{adjustbox}
\end{table}

\begin{table}[h]
\centering
\caption{Normalized validation error of each method on each benchmark for LLAMA3 8B Instruct. Averaged over repetitions. Standard errors in parentheses. For visual analysis, we highlight all methods that have a mean error that is less or equal to the mean error of the best method plus two times its standard error.}
\label{tab:raw_valid_llama3}
\centering
\begin{adjustbox}{width=1.00\linewidth,center}
\begin{tabular}{lrrrrrrrrrr}
\toprule
& \multicolumn{10}{c}{\textbf{Benchmark}}\\ \cline{2-11}
 & AI2 ARC & GSM8K & antonyms & larger animal & negation & object counting & orthography starts with & second word letter & sentiment & word unscrambling \\
\midrule
RS & 0.078 (0.010) & 0.067 (0.006) & 0.093 (0.010) & 0.023 (0.004) & 0.011 (0.001) & 0.156 (0.015) & 0.230 (0.031) & 0.275 (0.021) & 0.022 (0.003) & 0.271 (0.037) \\
vanilla BO & 0.051 (0.005) & 0.061 (0.007) & 0.079 (0.009) & 0.016 (0.003) & 0.010 (0.001) & 0.115 (0.017) & 0.184 (0.024) & 0.226 (0.023) & 0.024 (0.005) & 0.248 (0.036) \\
HDBO & 0.042 (0.006) & 0.019 (0.006) & 0.073 (0.009) & 0.015 (0.002) & \textbf{0.008 (0.001)} & \textbf{0.066 (0.008)} & 0.148 (0.023) & 0.186 (0.020) & \textbf{0.010 (0.003)} & 0.163 (0.027) \\
BOPCA & 0.049 (0.006) & 0.026 (0.007) & 0.074 (0.010) & 0.014 (0.002) & 0.010 (0.001) & \textbf{0.074 (0.015)} & 0.225 (0.028) & 0.160 (0.023) & \textbf{0.013 (0.004)} & 0.149 (0.023) \\
EASE & 0.059 (0.006) & 0.060 (0.005) & 0.115 (0.012) & 0.032 (0.005) & 0.010 (0.001) & \textbf{0.076 (0.005)} & 0.129 (0.022) & 0.271 (0.015) & 0.023 (0.004) & 0.357 (0.033) \\
MIPROv2 & 0.063 (0.007) & 0.056 (0.007) & 0.062 (0.008) & 0.017 (0.002) & \textbf{0.008 (0.001)} & 0.092 (0.011) & 0.183 (0.026) & 0.218 (0.022) & 0.023 (0.004) & 0.265 (0.033) \\
TRIPLE-SH & \textbf{0.023 (0.005)} & 0.007 (0.002) & \textbf{0.024 (0.005)} & \textbf{0.008 (0.002)} & 0.012 (0.002) & \textbf{0.068 (0.008)} & \textbf{0.034 (0.008)} & \textbf{0.047 (0.014)} & 0.025 (0.005) & 0.095 (0.019) \\
TRIPLE-GSE & \textbf{0.022 (0.006)} & \textbf{0.002 (0.001)} & \textbf{0.026 (0.006)} & 0.020 (0.003) & 0.015 (0.001) & \textbf{0.080 (0.008)} & 0.074 (0.011) & \textbf{0.071 (0.016)} & 0.035 (0.004) & 0.184 (0.033) \\
\hbbodkl & \textbf{0.019 (0.005)} & 0.006 (0.002) & \textbf{0.024 (0.004)} & \textbf{0.008 (0.002)} & \textbf{0.008 (0.001)} & 0.092 (0.018) & \textbf{0.041 (0.009)} & \textbf{0.043 (0.014)} & 0.020 (0.004) & \textbf{0.048 (0.014)} \\
\bottomrule
\end{tabular}
\end{adjustbox}
\end{table}

\begin{table}[h]
\centering
\caption{Normalized validation error of each method on each benchmark for Mistral 7B Instruct. Averaged over repetitions. Standard errors in parentheses. For visual analysis, we highlight all methods that have a mean error that is less or equal to the mean error of the best method plus two times its standard error.}
\label{tab:raw_valid_mistral7b}
\centering
\begin{adjustbox}{width=1.00\linewidth,center}
\begin{tabular}{lrrrrrrrrrr}
\toprule
& \multicolumn{10}{c}{\textbf{Benchmark}}\\ \cline{2-11}
 & AI2 ARC & GSM8K & antonyms & larger animal & negation & object counting & orthography starts with & second word letter & sentiment & word unscrambling \\
\midrule
RS & 0.329 (0.035) & 0.160 (0.021) & 0.020 (0.002) & 0.043 (0.006) & 0.006 (0.003) & 0.040 (0.004) & 0.164 (0.018) & 0.171 (0.016) & 0.098 (0.013) & 0.224 (0.017) \\
vanilla BO & 0.359 (0.038) & 0.166 (0.021) & 0.017 (0.002) & 0.040 (0.008) & 0.008 (0.003) & \textbf{0.033 (0.005)} & 0.117 (0.020) & 0.136 (0.013) & 0.085 (0.016) & 0.161 (0.019) \\
HDBO & 0.217 (0.045) & 0.105 (0.019) & \textbf{0.013 (0.002)} & 0.021 (0.005) & 0.009 (0.003) & \textbf{0.028 (0.005)} & 0.120 (0.018) & 0.085 (0.017) & \textbf{0.042 (0.013)} & 0.108 (0.016) \\
BOPCA & 0.246 (0.044) & 0.108 (0.019) & 0.015 (0.002) & 0.021 (0.006) & 0.011 (0.004) & \textbf{0.032 (0.006)} & 0.112 (0.021) & 0.125 (0.019) & \textbf{0.048 (0.013)} & 0.136 (0.021) \\
EASE & 0.384 (0.033) & 0.133 (0.014) & 0.022 (0.002) & 0.053 (0.008) & \textbf{0.000 (0.000)} & 0.035 (0.005) & 0.147 (0.023) & 0.183 (0.014) & 0.110 (0.014) & 0.095 (0.021) \\
MIPROv2 & 0.323 (0.041) & 0.122 (0.017) & \textbf{0.013 (0.002)} & 0.032 (0.005) & 0.011 (0.004) & \textbf{0.032 (0.005)} & 0.125 (0.014) & 0.158 (0.016) & 0.068 (0.014) & 0.163 (0.017) \\
TRIPLE-SH & \textbf{0.024 (0.011)} & 0.008 (0.003) & 0.041 (0.005) & \textbf{0.012 (0.003)} & 0.044 (0.008) & 0.057 (0.006) & 0.005 (0.004) & \textbf{0.044 (0.012)} & 0.055 (0.010) & 0.103 (0.018) \\
TRIPLE-GSE & \textbf{0.038 (0.016)} & 0.001 (0.000) & 0.042 (0.005) & 0.022 (0.006) & 0.055 (0.010) & 0.067 (0.010) & \textbf{0.002 (0.000)} & 0.072 (0.015) & 0.084 (0.017) & 0.145 (0.022) \\
\hbbodkl & 0.107 (0.036) & \textbf{0.000 (0.000)} & \textbf{0.011 (0.001)} & \textbf{0.012 (0.003)} & 0.012 (0.005) & \textbf{0.025 (0.005)} & 0.024 (0.010) & 0.067 (0.013) & \textbf{0.030 (0.011)} & \textbf{0.061 (0.014)} \\
\bottomrule
\end{tabular}
\end{adjustbox}
\end{table}

In Tables~\ref{tab:raw_valid_haiku}, \ref{tab:raw_valid_llama3} and \ref{tab:raw_valid_mistral7b} we report the average normalized validation error of the best prompt found by each method after having used a fraction of $1.00$ LLM calls, separately for each benchmark task, separately for each LLM.

We perform the same analysis for the test error at a fraction of $1.00$ and reject the global null hypothesis of no effect of methods ($\chi^2(8) = 288.36, p < 1\mathrm{e}{-4}$).
The pairwise results are:
{\tiny
\begin{itemize}
    \item \texttt{RS} vs. \hbbodkl, $z = 11.51, p < 1\mathrm{e}{-4}$
    \item vanilla \texttt{BO} vs. \hbbodkl, $z = 11.03, p < 1\mathrm{e}{-4}$
    \item \texttt{HDBO} vs. \hbbodkl, $z = 6.19, p < 1\mathrm{e}{-4}$
    \item \texttt{BOPCA} vs. \hbbodkl, $z = 7.58, p < 1\mathrm{e}{-4}$
    \item \texttt{EASE} vs. \hbbodkl, $z = 7.98, p < 1\mathrm{e}{-4}$
    \item \texttt{MIPROv2} vs. \hbbodkl, $z = 8.61, p < 1\mathrm{e}{-4}$
    \item \texttt{TRIPLE-SH} vs. \hbbodkl, $z = 1.49, p = 1.000$
    \item \texttt{TRIPLE-GSE} vs. \hbbodkl, $z = 1.34, p = 1.000$
\end{itemize}
}
Conclusions are largely consistent with the analysis with respect to validation error, however, while \hbbodkl improves over \texttt{TRIPLE-SH} and \texttt{TRIPLE-GSE} also with respect to test error, the effects are not strong enough to be considered statistically significant at the $\alpha = 0.01$ level.
For brevity, we do not include results for fractions of $0.25$ or $0.50$ here, where we observed similar results but \hbbodkl again more strongly outperforming the other methods.

\begin{table}[h]
\centering
\caption{Normalized test error of each method on each benchmark for Claude 3 Haiku. Averaged over repetitions. Standard errors in parentheses. For visual analysis, we highlight all methods that have a mean error that is less or equal to the mean error of the best method plus two times its standard error.}
\label{tab:raw_test_haiku}
\centering
\begin{adjustbox}{width=1.00\linewidth,center}
\begin{tabular}{lrrrrrrrrrr}
\toprule
& \multicolumn{10}{c}{\textbf{Benchmark}}\\ \cline{2-11}
 & AI2 ARC & GSM8K & antonyms & larger animal & negation & object counting & orthography starts with & second word letter & sentiment & word unscrambling \\
\midrule
RS & \textbf{0.224 (0.023)} & 0.092 (0.012) & \textbf{0.074 (0.003)} & 0.096 (0.008) & 0.124 (0.007) & 0.140 (0.014) & \textbf{0.277 (0.011)} & 0.058 (0.010) & 0.244 (0.019) & 0.163 (0.011) \\
vanilla BO & \textbf{0.213 (0.020)} & 0.092 (0.011) & 0.084 (0.004) & 0.113 (0.010) & 0.124 (0.007) & 0.147 (0.019) & 0.296 (0.011) & 0.063 (0.008) & \textbf{0.207 (0.016)} & 0.156 (0.012) \\
HDBO & \textbf{0.203 (0.020)} & 0.087 (0.010) & 0.078 (0.003) & 0.105 (0.008) & 0.124 (0.007) & \textbf{0.083 (0.015)} & 0.294 (0.011) & 0.030 (0.008) & \textbf{0.237 (0.014)} & 0.133 (0.009) \\
BOPCA & \textbf{0.209 (0.018)} & 0.095 (0.009) & 0.078 (0.003) & 0.105 (0.009) & 0.124 (0.007) & 0.129 (0.019) & 0.302 (0.012) & 0.039 (0.009) & \textbf{0.230 (0.014)} & 0.150 (0.011) \\
EASE & 0.243 (0.023) & \textbf{0.036 (0.006)} & \textbf{0.070 (0.004)} & 0.108 (0.007) & 0.124 (0.007) & 0.135 (0.015) & \textbf{0.282 (0.018)} & 0.058 (0.009) & 0.263 (0.016) & 0.135 (0.010) \\
MIPROv2 & \textbf{0.200 (0.020)} & 0.068 (0.006) & \textbf{0.077 (0.003)} & 0.099 (0.008) & 0.124 (0.007) & \textbf{0.100 (0.017)} & \textbf{0.271 (0.011)} & 0.039 (0.008) & \textbf{0.219 (0.016)} & 0.125 (0.008) \\
TRIPLE-SH & 0.256 (0.024) & 0.099 (0.009) & 0.078 (0.006) & \textbf{0.076 (0.006)} & \textbf{0.114 (0.011)} & \textbf{0.084 (0.012)} & \textbf{0.272 (0.015)} & \textbf{0.008 (0.005)} & 0.248 (0.019) & \textbf{0.116 (0.005)} \\
TRIPLE-GSE & 0.307 (0.024) & 0.080 (0.010) & 0.082 (0.005) & 0.115 (0.010) & \textbf{0.106 (0.009)} & 0.130 (0.019) & 0.299 (0.014) & \textbf{0.009 (0.005)} & 0.322 (0.022) & \textbf{0.122 (0.005)} \\
\hbbodkl & \textbf{0.182 (0.026)} & 0.093 (0.008) & 0.080 (0.003) & 0.089 (0.007) & \textbf{0.118 (0.008)} & 0.117 (0.021) & \textbf{0.262 (0.010)} & \textbf{0.018 (0.007)} & \textbf{0.219 (0.016)} & \textbf{0.125 (0.004)} \\
\bottomrule
\end{tabular}
\end{adjustbox}
\end{table}

\begin{table}[h]
\centering
\caption{Normalized test error of each method on each benchmark for LLAMA3 8B Instruct. Averaged over repetitions. Standard errors in parentheses. For visual analysis, we highlight all methods that have a mean error that is less or equal to the mean error of the best method plus two times its standard error.}
\label{tab:raw_test_llama3}
\centering
\begin{adjustbox}{width=1.00\linewidth,center}
\begin{tabular}{lrrrrrrrrrr}
\toprule
& \multicolumn{10}{c}{\textbf{Benchmark}}\\ \cline{2-11}
 & AI2 ARC & GSM8K & antonyms & larger animal & negation & object counting & orthography starts with & second word letter & sentiment & word unscrambling \\
\midrule
RS & 0.066 (0.010) & 0.112 (0.011) & 0.147 (0.012) & 0.085 (0.011) & \textbf{0.104 (0.010)} & 0.238 (0.021) & 0.296 (0.041) & 0.270 (0.029) & 0.356 (0.013) & 0.633 (0.054) \\
vanilla BO & 0.038 (0.009) & 0.099 (0.012) & 0.145 (0.013) & \textbf{0.075 (0.007)} & 0.129 (0.008) & 0.236 (0.023) & 0.218 (0.034) & 0.292 (0.031) & 0.356 (0.015) & 0.700 (0.054) \\
HDBO & 0.030 (0.008) & 0.028 (0.009) & 0.133 (0.014) & 0.086 (0.007) & 0.138 (0.007) & 0.165 (0.019) & 0.184 (0.035) & 0.248 (0.027) & 0.321 (0.010) & 0.767 (0.056) \\
BOPCA & 0.033 (0.004) & 0.038 (0.011) & 0.124 (0.014) & \textbf{0.075 (0.007)} & 0.126 (0.009) & 0.216 (0.014) & 0.283 (0.038) & 0.227 (0.025) & 0.344 (0.010) & 0.667 (0.048) \\
EASE & 0.047 (0.007) & 0.094 (0.009) & 0.181 (0.015) & 0.076 (0.009) & \textbf{0.112 (0.008)} & \textbf{0.014 (0.009)} & 0.171 (0.037) & 0.248 (0.027) & \textbf{0.274 (0.013)} & 0.622 (0.055) \\
MIPROv2 & 0.041 (0.008) & 0.088 (0.012) & 0.114 (0.013) & 0.076 (0.006) & 0.135 (0.009) & 0.160 (0.019) & 0.239 (0.040) & 0.248 (0.024) & 0.356 (0.014) & 0.700 (0.051) \\
TRIPLE-SH & \textbf{0.016 (0.003)} & 0.008 (0.003) & \textbf{0.070 (0.015)} & \textbf{0.075 (0.008)} & 0.125 (0.011) & 0.177 (0.017) & \textbf{0.037 (0.009)} & 0.152 (0.011) & 0.362 (0.014) & 0.678 (0.059) \\
TRIPLE-GSE & \textbf{0.017 (0.004)} & \textbf{0.003 (0.002)} & \textbf{0.052 (0.012)} & \textbf{0.059 (0.008)} & \textbf{0.110 (0.009)} & 0.167 (0.019) & 0.078 (0.019) & \textbf{0.140 (0.016)} & 0.369 (0.014) & \textbf{0.544 (0.059)} \\
\hbbodkl & \textbf{0.016 (0.006)} & 0.009 (0.003) & 0.093 (0.015) & \textbf{0.064 (0.007)} & 0.152 (0.010) & 0.207 (0.027) & 0.078 (0.019) & \textbf{0.120 (0.011)} & 0.341 (0.012) & \textbf{0.489 (0.044)} \\
\bottomrule
\end{tabular}
\end{adjustbox}
\end{table}

\begin{table}[h]
\centering
\caption{Normalized test error of each method on each benchmark for Mistral 7B Instruct. Averaged over repetitions. Standard errors in parentheses. For visual analysis, we highlight all methods that have a mean error that is less or equal to the mean error of the best method plus two times its standard error.}
\label{tab:raw_test_mistral7b}
\centering
\begin{adjustbox}{width=1.00\linewidth,center}
\begin{tabular}{lrrrrrrrrrr}
\toprule
& \multicolumn{10}{c}{\textbf{Benchmark}}\\ \cline{2-11}
 & AI2 ARC & GSM8K & antonyms & larger animal & negation & object counting & orthography starts with & second word letter & sentiment & word unscrambling \\
\midrule
RS & 0.277 (0.037) & 0.198 (0.021) & 0.099 (0.006) & 0.239 (0.033) & 0.502 (0.025) & \textbf{0.188 (0.015)} & 0.293 (0.026) & 0.194 (0.015) & 0.156 (0.019) & 0.476 (0.039) \\
vanilla BO & 0.366 (0.042) & 0.200 (0.022) & \textbf{0.092 (0.006)} & 0.201 (0.025) & 0.531 (0.025) & \textbf{0.199 (0.013)} & 0.229 (0.026) & 0.172 (0.018) & 0.125 (0.017) & 0.448 (0.031) \\
HDBO & 0.215 (0.045) & 0.112 (0.020) & 0.098 (0.005) & \textbf{0.140 (0.009)} & \textbf{0.481 (0.027)} & 0.226 (0.013) & 0.237 (0.031) & 0.112 (0.021) & \textbf{0.059 (0.018)} & 0.386 (0.028) \\
BOPCA & 0.266 (0.047) & 0.125 (0.019) & \textbf{0.088 (0.007)} & 0.182 (0.019) & 0.500 (0.028) & 0.232 (0.013) & 0.226 (0.031) & 0.140 (0.022) & \textbf{0.063 (0.016)} & 0.357 (0.034) \\
EASE & 0.356 (0.036) & 0.174 (0.014) & \textbf{0.076 (0.008)} & 0.181 (0.021) & 0.631 (0.022) & \textbf{0.209 (0.013)} & 0.257 (0.029) & 0.211 (0.018) & 0.158 (0.019) & \textbf{0.290 (0.025)} \\
MIPROv2 & 0.327 (0.043) & 0.165 (0.016) & 0.094 (0.004) & 0.178 (0.020) & 0.519 (0.026) & \textbf{0.195 (0.013)} & 0.248 (0.025) & 0.193 (0.018) & 0.115 (0.022) & 0.429 (0.029) \\
TRIPLE-SH & \textbf{0.016 (0.008)} & 0.078 (0.010) & 0.104 (0.007) & \textbf{0.140 (0.009)} & 0.526 (0.026) & 0.230 (0.014) & \textbf{0.044 (0.014)} & \textbf{0.061 (0.016)} & 0.116 (0.019) & 0.395 (0.033) \\
TRIPLE-GSE & \textbf{0.028 (0.012)} & 0.050 (0.011) & 0.095 (0.006) & 0.169 (0.011) & \textbf{0.421 (0.038)} & 0.226 (0.013) & \textbf{0.061 (0.011)} & \textbf{0.071 (0.015)} & 0.104 (0.022) & 0.400 (0.029) \\
\hbbodkl & 0.096 (0.032) & \textbf{0.011 (0.006)} & \textbf{0.083 (0.005)} & 0.162 (0.013) & \textbf{0.462 (0.028)} & 0.238 (0.011) & \textbf{0.053 (0.023)} & \textbf{0.083 (0.017)} & \textbf{0.088 (0.021)} & 0.367 (0.030) \\
\bottomrule
\end{tabular}
\end{adjustbox}
\end{table}

In Tables~\ref{tab:raw_test_haiku}, \ref{tab:raw_test_llama3} and \ref{tab:raw_test_mistral7b} we report the average normalized test error of the best prompt found by each method after having used a fraction of $1.00$ LLM calls, separately for each benchmark task, separately for each LLM.

\subsection{Ablation Study}\label{app:additional_results_ablation}
Here, we provide additional analyses of the ablation results reported in Section~\ref{sec:ablation_results}.
To test the significance of each component on the validation and test error of \hbbodkl, we again conduct a linear mixed effects model analysis.
We model the unaggregated performance as a function of the fraction of LLM calls (i.e., over time; starting after the initial design of full-fidelity methods, i.e., after a fraction of $0.40$) and include random intercepts for each benchmark scenario (benchmark task and LLM combination).
To test the global hypothesis that there is an effect of the components on performance, we test an intercept model against a model including an effect of the factor method (corresponding to an ablation variant).
If we reject the null hypothesis, we proceed with a Tukey post hoc test (corrected for multiple testing) to perform pairwise comparisons.
We test at the conservative $\alpha = 0.01$ level.

Examining the validation error, we reject the global null hypothesis of no effect of the method on the anytime performance ($\chi^2(5) = 35184.28, p < 1\mathrm{e}{-4}$).
The relevant pairwise comparison results state as follows:
{\tiny
\begin{itemize}
    \item vanilla \texttt{BO} vs. \texttt{BoPs (non structural-aware DK-GP)}, $z = 48.80, p < 1\mathrm{e}{-4}$
    \item \texttt{BoPs (non structural-aware DK-GP)} vs. \texttt{BoPs (structural-aware DK-GP)}, $z = 16.67, p < 1\mathrm{e}{-4}$
    \item \texttt{BoPs (structural-aware DK-GP)} vs. \texttt{HB}, $z = 52.92, p < 1\mathrm{e}{-4}$
    \item \texttt{HB} vs. \hbbodkl, $z = 24.00, p < 1\mathrm{e}{-4}$
\end{itemize}
}
We can conclude that using a DK-GP significantly improves over vanilla BO, that a structural-aware DK-GP improves over the non structural-aware DK-GP, that HB improves over the structural-aware DK-GP and that \hbbodkl further improves over HB.

We perform the same analysis for the test error and reject the global hypothesis of no effect ($\chi^2(5) = 14578.73, p < 1\mathrm{e}{-4}$).
The relevant pairwise comparison results state as follows:
{\tiny
\begin{itemize}
    \item vanilla \texttt{BO} vs. \texttt{BoPs (non structural-aware DK-GP)}, $z = 39.35, p < 1\mathrm{e}{-4}$
    \item \texttt{BoPs (non structural-aware DK-GP)} vs. \texttt{BoPs (structural-aware DK-GP)}, $z = 6.51, p < 1\mathrm{e}{-4}$
    \item \texttt{BoPs (structural-aware DK-GP)} vs. \texttt{HB}, $z = 31.59, p < 1\mathrm{e}{-4}$
    \item \texttt{HB} vs. \hbbodkl, $z = 18.85, p < 1\mathrm{e}{-4}$
\end{itemize}
}
Conclusions are the same as for the validation error.

\subsection{Encoder Sensitivity}\label{app:additional_results_embedding_sensitivity}
Here, we provide additional analyses of the encoder sensitivity results reported in Section~\ref{sec:embedding_sensitivity}.
To test whether the choice of encoder model affects the final performance of \hbbodkl, we again conduct a linear mixed effects model analysis.
We model the unaggregated performance at a fraction of $1.00$ total LLM calls involving random intercepts for each benchmark scenario (benchmark task and LLM combination).
To test the global hypothesis that there is an effect of the encoder on performance, we test an intercept model against a model including an effect of the factor encoder.
We test at the conservative $\alpha = 0.01$ level.
For both the validation and test error, we cannot reject the null hypothesis of the encoder making no difference, $\chi^2(2) = 4.69, p = 0.096$ and $\chi^2(2) = 3.85, p = 0.146$ respectively.
We therefore conclude that \hbbodkl is robust to the choice of encoder model.

\subsection{Hyperband Design Choices}\label{app:additional_results_hb_ablation}
\begin{figure}[h]
    \centering
    \begin{subfigure}{.5\textwidth}
        \includegraphics[width=\linewidth]{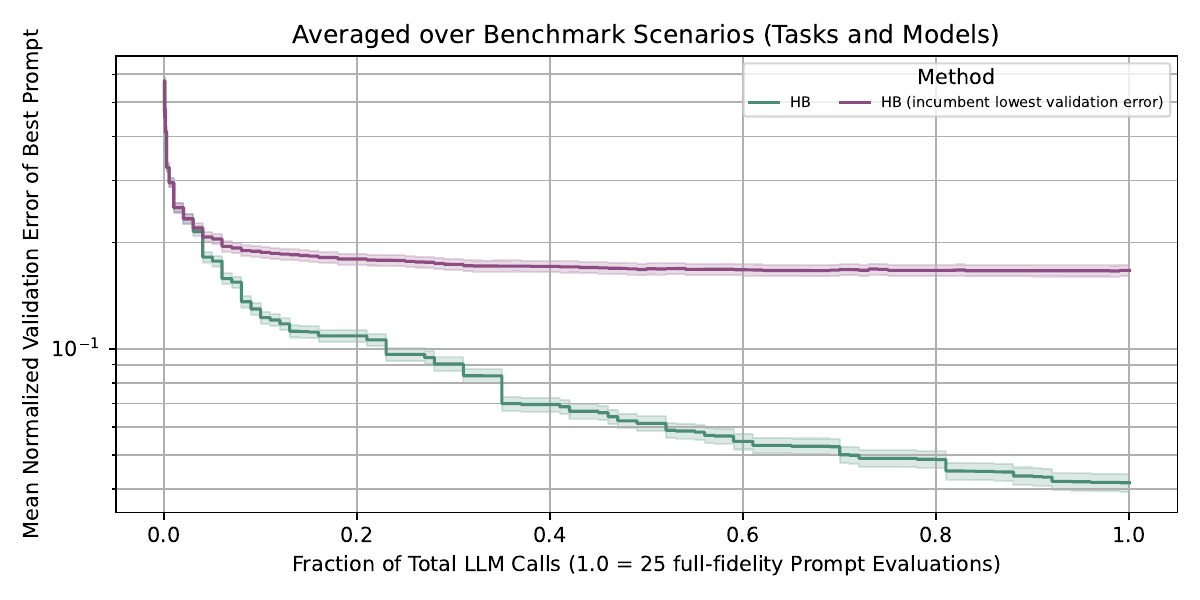}
        \caption{Validation}
        \label{fig:hb_ablation_selection_overall_valid}
        \vspace{-1em}
    \end{subfigure}%
    \begin{subfigure}{.5\textwidth}
        \centering
        \includegraphics[width=\linewidth]{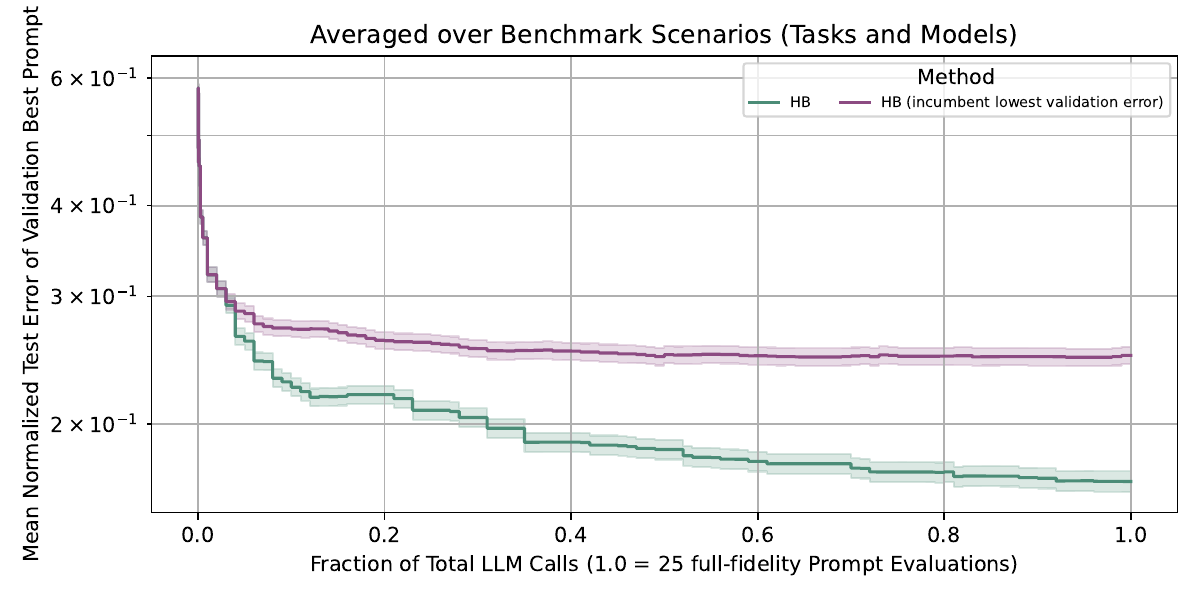}
        \caption{Test}
        \label{fig:hb_ablation_selection_overall_test}
        \vspace{-1em}
    \end{subfigure}%
    \caption{Normalized error (log scale) of the best prompt found by each HB incumbent selection mechanism, averaged over benchmarks. Lower is better. Ribbons represent SE.}
    \label{fig:hb_ablation_selection_overall}
\end{figure}

As mentioned in Section~\ref{sec:method} and Appendix~\ref{app:mf}, adapting HB to prompt selection involves several design decisions.
Here, we provide an ablation of these decisions.
While vanilla HB for hyperparameter optimization would return the configuration with the lowest validation error as the (anytime) incumbent, this is not sensible for prompt selection as the fidelity directly influences the noise of the validation error.
We therefore always return the prompt with the lowest validation error among all prompts that have been evaluated on the (current) highest fidelity level.
To analyze the effect of this design decision, we run HB for prompt selection with this incumbent selection mechanism and compare to the incumbent selection mechanism that simply selects the prompt with the lowest validation error.
The experimental setup is exactly the same as for the results reported in the main paper.
We visualize the (oracle) normalized validation and test error of the best prompt found by HB under each incumbent selection scheme in Figure~\ref{fig:hb_ablation_selection_overall}.
As before, for visualization purposes the validation error of the incumbent is computed here in an oracle setting (i.e., using all validation instances), whereas during the selection process the anytime incumbent itself was selected based on its validation error computed on fewer validation instances.
Examining the validation error (Figure~\ref{fig:hb_ablation_selection_overall_valid}) of \texttt{HB} as used by us (selecting the incumbent as the prompt with the lowest validation error among all prompts evaluated on the highest fidelity level), we observe that the validation error of the incumbent keeps decreasing as optimization progresses.
In contrast, if we would perform the incumbent selection simply by choosing the prompt with the lowest validation error (ignoring the fidelity level), as in \texttt{HB (incumbent lowest validation error)}, we observe that optimization progress stagnates quickly.
This occurs because the incumbent is no longer updated, since, at lower fidelity levels, noisy performance estimates can result in artificially low validation errors.
Similar conclusions hold for the test error (Figure~\ref{fig:hb_ablation_selection_overall_test}).

To test whether the choice of selection mechanism does make a difference for the final performance of HB, we again conduct a linear mixed effects model analysis.
We model the unaggregated performance at a fraction of $1.00$ total LLM calls involving random intercepts for each benchmark scenario (benchmark task and LLM combination).
To test the global hypothesis that there is an effect of the incumbent selection mechanism on performance, we test an intercept model against a model including an effect of the factor selection mechanism.
We test at the conservative $\alpha = 0.01$ level.
For both validation and test error we reject the null hypothesis of no effect of the selection mechanism $\chi^2(1) = 448.05, p < 1\mathrm{e}{-4}$ and $\chi^2(1) = 164.17, p < 1\mathrm{e}{-4}$ respectively.
We therefore conclude that our incumbent selection mechanism is superior.

Another design decision for adapting HB to prompt selection is concerned with whether prompts should be evaluated on the same random validation instances within a stage or on their own random samples.
The final related design decision involves whether validation instances of higher stages for a given bracket are constructed to be supersets of the validation instances used in lower stages (as described in Appendix~\ref{app:mf}) which allows for further speed-ups due to caching.

\begin{figure}[h]
    \centering
    \begin{subfigure}{.5\textwidth}
        \includegraphics[width=\linewidth]{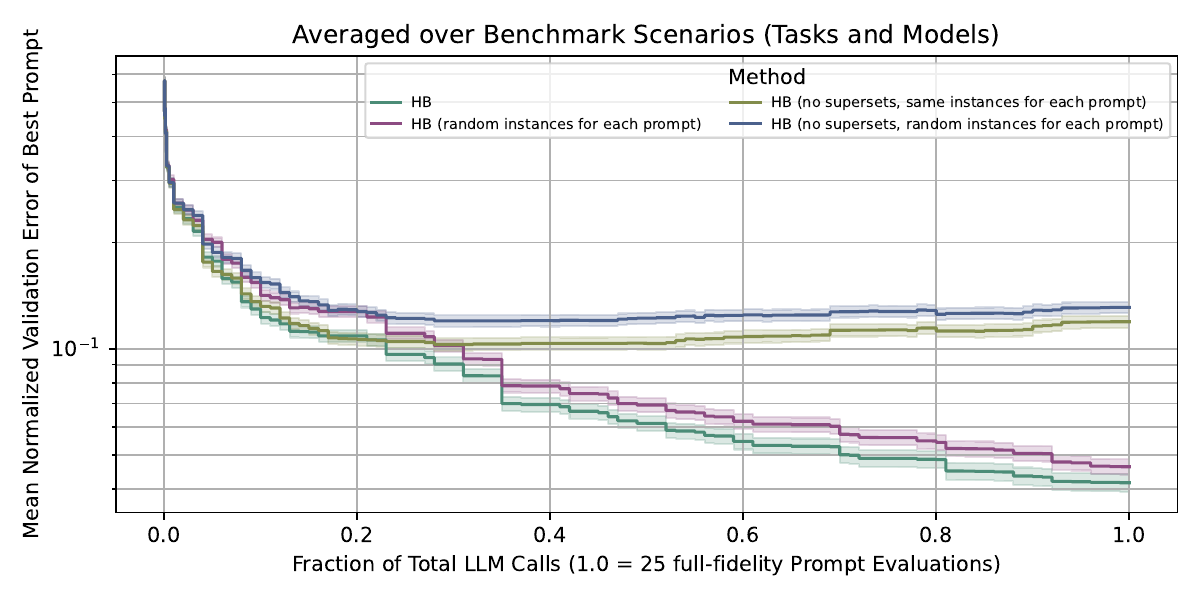}
        \caption{Validation}
        \label{fig:hb_ablation_overall_valid}
        \vspace{-1em}
    \end{subfigure}%
    \begin{subfigure}{.5\textwidth}
        \centering
        \includegraphics[width=\linewidth]{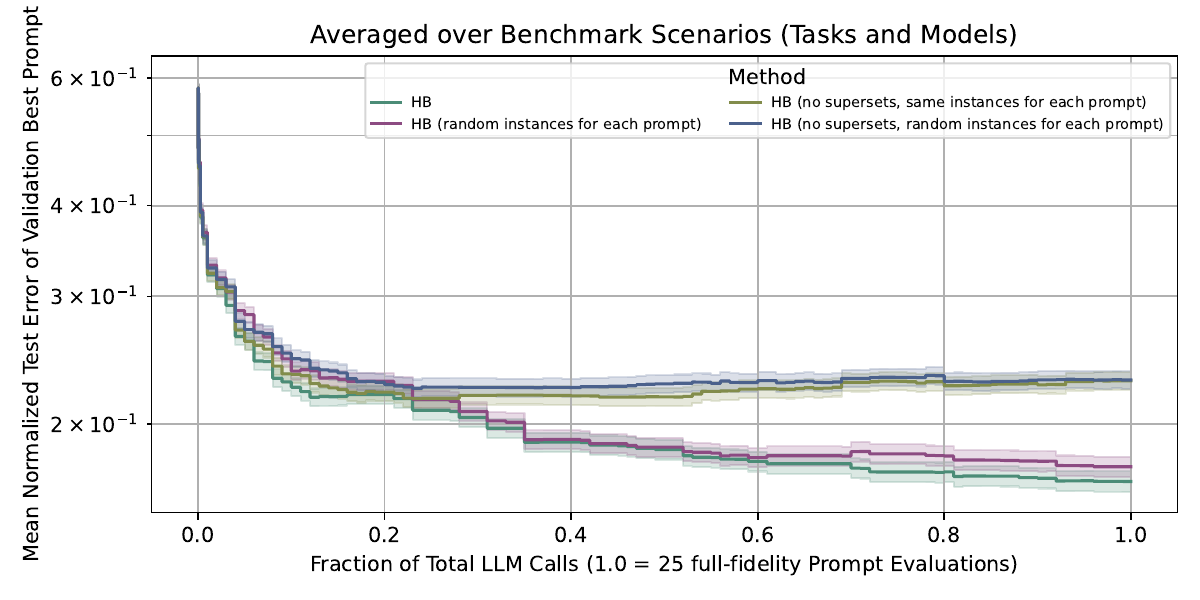}
        \caption{Test}
        \label{fig:hb_ablation_overall_test}
        \vspace{-1em}
    \end{subfigure}%
    \caption{Normalized error (log scale) of the best prompt found by each HB validation instances sampling variant, averaged over benchmarks. Lower is better. Ribbons represent SE.}
    \label{fig:hb_ablation_overall}
\end{figure}

To investigate the effect of using the same random vs. truly random instances for each prompt and the effect of validation instances used in higher stages of a bracket being supersets of the validation instances used in lower stages, we run HB for prompt selection varying these two components.
The experimental setup is exactly the same as for the results reported in the main paper.
We visualize the (oracle) normalized validation and test error of the best prompt found by HB under each incumbent selection scheme in Figure~\ref{fig:hb_ablation_overall}.
As before, for visualization purposes the validation error of the incumbent is computed here in an oracle setting (i.e., using all validation instances), whereas during the selection process the anytime incumbent itself was selected based on its validation error computed on fewer validation instances.
Examining the validation error (Figure~\ref{fig:hb_ablation_overall_valid}) of our proposed \texttt{HB} (same random instances and supersets), we can see that this variant performs best.
If we use truly random instances (but keep the superset structure) as in \texttt{HB (random instances for each prompt)}, performance is slightly worse.
Giving up the superset structure (\texttt{HB (no supersets, same instances for each prompt)} and \texttt{HB (no supersets, random instances for each prompt)} we can see that performance is substantially worse, even more so when using truly random instances for each prompt.
In general, we can conclude that the effect of using supersets for higher stages within a given bracket boosts the performance of HB.
Moreover, using truly random validation instances for each prompt instead of using the same random validation instances for all prompt evaluations in a stage (i.e., the paired setting) generally worsens performance.
Examining the test performance, we observe that these conclusions generalize (Figure~\ref{fig:hb_ablation_overall_test}).

To test whether the design decisions of using supersets and using the same random validation instances for each prompt evaluation within a stage make a difference for the final performance of HB, we again conduct a linear mixed effects model analysis.
We model the unaggregated performance at a fraction of $1.00$ total LLM calls involving random intercepts for each benchmark scenario (benchmark task and LLM combination).
To test the global hypothesis that there is an effect of the design choices on performance, we test an intercept model against a model including an effect of the factor method (corresponding to an ablation variant).
We test at the conservative $\alpha = 0.01$ level.

For the validation error at a fraction of $1.00$, we reject the global null hypothesis of no effect of methods ($\chi^2(3) = 730.85, p < 1\mathrm{e}{-4}$).
The pairwise results are:
{\tiny
\begin{itemize}
\item \texttt{HB (random instances for each prompt)} vs. \texttt{HB}, $z = 1.13, p = 0.259$
\item \texttt{HB (no supersets, same instances for each prompt)} vs. \texttt{HB}, $z = 19.12, p < 1\mathrm{e}{-4}$
\item \texttt{HB (no supersets, random instances for each prompt)} vs. \texttt{HB}, $z = 22.03, p < 1\mathrm{e}{-4}$
\item \texttt{HB (no supersets, same instances for each prompt)} vs. \texttt{HB (random instances for each prompt)}, $z = 17.99, p < 1\mathrm{e}{-4}$
\item \texttt{HB (no supersets, random instances for each prompt)} vs. \texttt{HB (random instances for each prompt)}, $z = 20.90, p < 1\mathrm{e}{-4}$
\item \texttt{HB (no supersets, random instances for each prompt)} vs. \texttt{HB (no supersets, same instances for each prompt)}, $z = 2.91, p = 0.007$
\end{itemize}
}

Examining test error at a fraction of $1.00$ we also reject the global null hypothesis of no effect of methods ($\chi^2(3) = 207.12, p < 1\mathrm{e}{-4}$).
The pairwise results are:
{\tiny
\begin{itemize}
\item \texttt{HB (random instances for each prompt)} vs. \texttt{HB}, $z = 1.39, p = 0.328$
\item \texttt{HB (no supersets, same instances for each prompt)} vs. \texttt{HB}, $z = 10.92, p < 1\mathrm{e}{-4}$
\item \texttt{HB (no supersets, random instances for each prompt)} vs. \texttt{HB}, $z = 11.02, p < 1\mathrm{e}{-4}$
\item \texttt{HB (no supersets, same instances for each prompt)} vs. \texttt{HB (random instances for each prompt)}, $z = 9.53, p < 1\mathrm{e}{-4}$
\item \texttt{HB (no supersets, random instances for each prompt)} vs. \texttt{HB (random instances for each prompt)}, $z = 9.63, p < 1\mathrm{e}{-4}$
\item \texttt{HB (no supersets, random instances for each prompt)} vs. \texttt{HB (no supersets, same instances for each prompt)}, $z = 0.11, p = 0.916$
\end{itemize}
}

Summarizing, our results confirm that the design decisions made to adapt HB to prompt selection are effective:
(1) The incumbent should be selected as the prompt with the lowest validation error among all prompts evaluated on the highest fidelity level.
(2) Validation instances used to evaluate prompts in higher stages of a given bracket should be supersets of the validation instances used in lower stages.
(3) Using the same (random) validation instances to evaluate prompts in each stage in general is beneficial compared to using truly random validation instances for each prompt, albeit this effect is comparably small.

\end{document}